\documentclass[acmtog]{acmart}

\usepackage{booktabs} %

\usepackage{multirow}
\usepackage{makecell}
\usepackage{amsmath}

\usepackage{colortbl}
\usepackage{graphicx}
\usepackage{subcaption}
\usepackage{lipsum}
\usepackage{arydshln}

\makeatletter
\DeclareRobustCommand\onedot{\futurelet\@let@token\@onedot}
\def\@onedot{\ifx\@let@token.\else.\null\fi\xspace}

\makeatother

\definecolor{yellow}{rgb}{1, 1, 0.7}
\definecolor{orange}{rgb}{1, 0.85, 0.7}
\definecolor{tablered}{rgb}{1, 0.7, 0.7}
\definecolor{red}{rgb}{1, 0, 0}

\definecolor{wincolor}{rgb}{0.85, 0.0, 0.0}

\definecolor{darkyellow}{rgb}{0.8, 0.8, 0.5}
\definecolor{darkred}{rgb}{0.7, 0.3, 0.3}
\definecolor{darkgreen}{rgb}{0.3, 0.7, 0.3}
\definecolor{green}{rgb}{0, 1.0, 0}
\definecolor{pink}{rgb}{1, 0.4, 0.7}

\definecolor{realred}{rgb}{0.95, 0.1, 0.0}

\usepackage{wrapfig}
\usepackage{colortbl}
\usepackage{subcaption}

\usepackage{multirow}
\usepackage{booktabs} %
\usepackage{color}
\definecolor{applegreen}{rgb}{0.55, 0.71, 0.0}
\definecolor{autumnorange}{rgb}{0.87, 0.61, 0.33}
\definecolor{moreprompt}{HTML}{C2D9FF}
\definecolor{oneprompt}{HTML}{FFE4D6}

\definecolor{heatYellow}{HTML}{FFFFB2}   %
\definecolor{heatOrange}{HTML}{FFD9B2}   %
\definecolor{heatRed}{HTML}{FFB2B2}   %

\definecolor{BrickRed}{HTML}{B5341E}
\definecolor{ForestGreen}{HTML}{309C59}

\definecolor{myPink}{HTML}{E91E63}  %

\usepackage{algorithm}
\usepackage{algpseudocode}
\usepackage{enumitem}

\usepackage{pifont}        %

\newcommand{\cmark}{\textcolor{ForestGreen}{\ding{52}}} %
\newcommand{\xmark}{\textcolor{BrickRed}{\ding{56}}}  

\usepackage[font=small,labelfont=bf]{caption}
\usepackage{subcaption}

\begin{document}

\copyrightyear{2025}
\acmYear{2025}
\acmConference[SA Conference Papers '25]{SIGGRAPH Asia 2025 Conference Papers}{December 15--18, 2025}{Hong Kong, Hong Kong}
\acmBooktitle{SIGGRAPH Asia 2025 Conference Papers (SA Conference Papers '25), December 15--18, 2025, Hong Kong, Hong Kong}\acmDOI{10.1145/3757377.3763815}
\acmISBN{979-8-4007-2137-3/2025/12}

\title{InfiniHuman: Infinite 3D Human Creation with Precise Control}

\author{Yuxuan Xue}
\affiliation{
\institution{University of Tübingen, Tübingen AI Center}
\country{Germany}
}

\author{Xianghui Xie}
\affiliation{
\institution{University of Tübingen, Tübingen AI Center, MPI for Informatics, SIC}
\country{Germany}
}

\author{Margaret Kostyrko}
\affiliation{
\institution{University of Tübingen}
\country{Germany}
}

\author{Gerard Pons-Moll}
\affiliation{
\institution{University of Tübingen, Tübingen AI Center, MPI for Informatics, SIC}
\country{Germany}
}

\begin{abstract}

Generating realistic and controllable 3D human avatars is a long-standing challenge. The difficulty increases when covering a broad range of attributes such as ethnicity, age, clothing styles, and detailed body shapes. 
Capturing and annotating large-scale human datasets for training generative models is prohibitively expensive and limited in both scale and diversity.
The central question we address in this paper is: \textit{Can we distill existing foundation models to generate theoretically unbounded richly annotated 3D human data?} 
We introduce \textbf{InfiniHuman}, a novel framework to distill these models synergistically, to generate richly annotated human data with minimal cost and theoretically unlimited scalability.
Specifically, we propose \textbf{InfiniHumanData}, a fully automatic pipeline that leverages vision-language and image generation models to create a large-scale multi-modal dataset. 
Remarkably, users cannot distinguish our automatically generated identities from scan renderings. 
InfiniHumanData contains \textbf{111K identities} and covers unprecedented diversity in ethnicity, age, clothing styles, and more. Each identity is annotated with multi-granularity text descriptions, multi-view RGB images, detailed clothing images, and SMPL body shape parameters.
Based on this, we learn \textbf{InfiniHumanGen}, a diffusion-based generative pipeline conditioned on text, body shape, and clothing assets. InfiniHumanGen enables fast, realistic, and precisely controllable avatar generation. 
Extensive experiments demonstrate that InfiniHuman significantly surpasses existing state-of-the-art methods in terms of visual quality, generation speed, and controllability.
Importantly, our approach democratizes high-quality avatar generation with fine-grained control at infinite scale through a practical and affordable solution.
To facilitate future research, we will publicly release our automatic data generation pipeline and the comprehensive dataset \textbf{InfiniHumanData}, and the generative models \textbf{InfiniHumanGen}.
The code and data of InfiniHuman is publicly available at \url{https://yuxuan-xue.com/infini-human}.
\end{abstract}



\begin{teaserfigure}
\begin{flushleft}
    \centering
    \includegraphics[width=\linewidth]{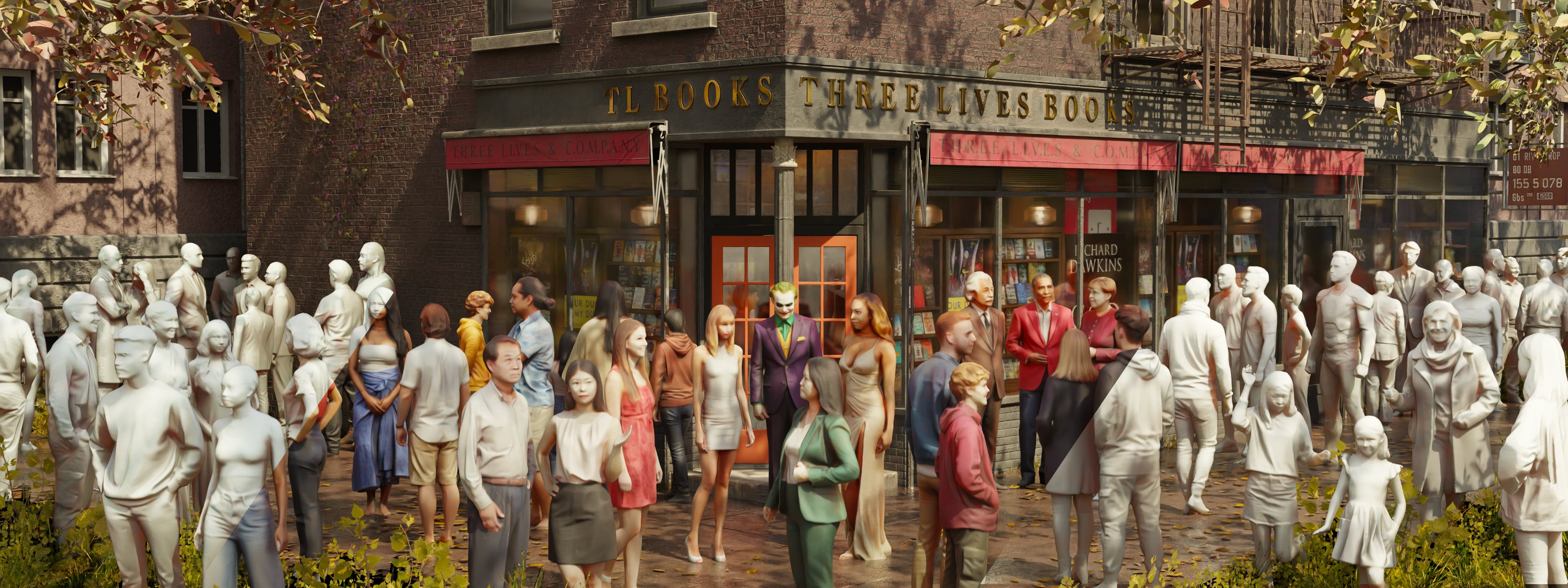}
    \caption{Using \textit{text description}, \textit{explicit body shape}, \textit{cloth image} as input, our 3D human generative method, \textbf{InfiniHuman}, can automatically create a variety of realistic 3D humans with high-fidelity texture and geometry. Our InfiniHuman allows for generating infinite 3D humans with precise user control.}
    \label{fig:teaser}
\end{flushleft}
\end{teaserfigure}

\maketitle

\section{Introduction}

\begin{figure*}
    \centering
    \includegraphics[width=\linewidth]{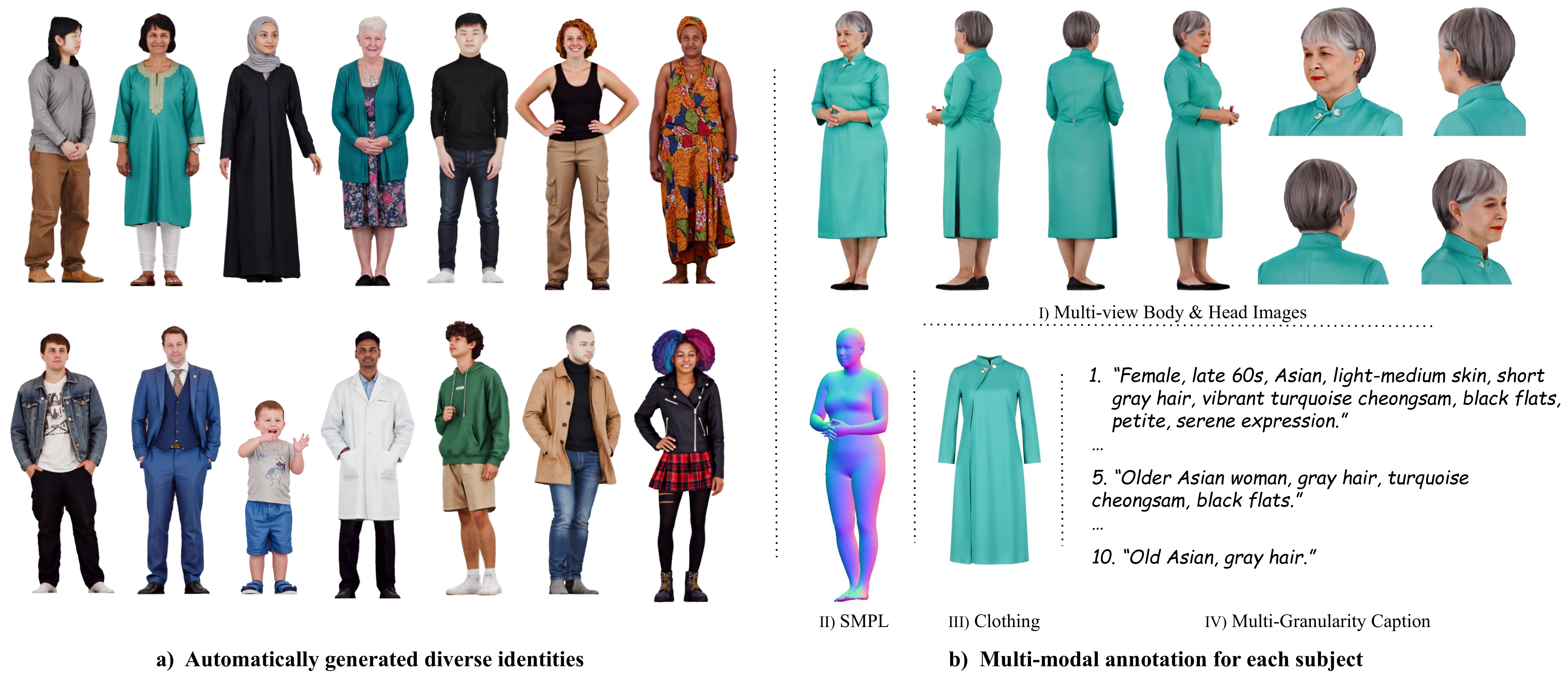}
    \caption{\textbf{Examples from InfiniHumanData}. \textbf{a)} Diverse human identities covering a wide range of ethnicities, age groups (including children), clothing styles, hair types, and skin tones, which are visually indistinguishable from real scans rendering (Sec.~\ref{sec:evaluation}). \textbf{b)} Multi-modal annotations per each subject, including I) multi-view RGB images (full-body and head), II) SMPL parameters, III) clothing asset images, and IV) multi-granularity text descriptions.}
    \label{fig:infinidata_examples}
\end{figure*}

Creating realistic and controllable 3D human avatars is a fundamental problem of growing significance in virtual reality, digital fashion, gaming, and social telepresence. Applications increasingly demand photorealistic avatars that can be personalized to match textual descriptions, specific body shapes, and user-provided clothing. However, the limitations of existing generation techniques have become increasingly apparent. In particular, generating diverse and semantically rich 3D humans, varying in clothing, ethnicity, age, gender, and shape, remains difficult due to the high cost and limited diversity of manually captured datasets.

Recent training-free approaches such as Score Distillation Sampling (SDS)~\cite{poole2023dreamfusion} have leveraged powerful text-to-image diffusion models to bypass dataset acquisition. However, these methods suffer from long optimization times, limited visual fidelity, and a lack of precise control over attributes like garment appearance or detailed body shape. These limitations motivate a critical research question: \emph{Can we distill the capabilities of foundation models to generate richly annotated 3D human data at theoretically unlimited scale and with precise controllability?}

We propose \textbf{InfiniHuman}, a fully automated framework that addresses this question by systematically repurposing and integrating existing vision-language, image synthesis, pose estimation, and diffusion models. Our method produces realistic 3D human identities at unprecedented scale, each annotated with multi-view images, fine-grained textual descriptions, SMPL parameters, and explicit clothing representations. The resulting dataset, \textbf{InfiniHumanData}, contains over 111K identities and supports detailed control across age, ethnicity, clothing, and body morphology.

Built upon this dataset, we introduce \textbf{InfiniHumanGen}, a pair of generative models capable of synthesizing 3D avatars conditioned jointly on text, clothing image and body shape, giving the user powerful controls. It includes two complementary models: \textbf{Gen-Schnell}, which enables rapid 3D generation and produces a Gaussian splatting output, and \textbf{Gen-HRes}, which produces high-resolution, photorealistic textured meshes. Our models outperform prior works on visual quality, speed, and attribute controllability, achieving state-of-the-art results with significantly lower computational cost.

In summary, the main technical contributions of our work include:
\begin{itemize}
    \item\textbf{InfiniHuman}, a framework to generate virtually unlimited richly annotated data of humans by distilling existing foundation models. The framework is fully automatic and generates identities indistinguishable from real scans. 
    \item\textbf{InfiniHumanData}, the first large-scale multi-modal human dataset comprising 111K diverse identities with rich multi-modal annotations essential for precise avatar generation.
    \item \textbf{InfiniHumanGen}, a novel generative framework supporting two distinct models: Gen-Schnell for fast and interactive 3D human generation and Gen-HRes for high-resolution and visually detailed 3D human creation; both from various user-specified inputs such as text, clothing, or body shape. 
\end{itemize}

By removing the need for costly scans, our method democratizes high-quality avatar creation, empowering applications in fashion, gaming, AR/VR, and beyond.

\section{Related work}

\subsection{3D Human Generation.} 
The creation of 3D human avatars from user-defined conditions is a long-standing problem in vision and graphics, with most prior works falling into two categories: reconstruction from images~\cite{saito2019pifu, xiu2022icon, xiu2023econ, zheng2021pamir, liao2025soap}, and generation from text~\cite{kim2022clipactor, han2023headsculpt, liao2023tada, yuan2024gavatar, hong2022avatarclip, cao2023dreamavatar, kolotouros2023dreamhuman, zhang2023avatarverse, wang2024disentangledavatar, liu2024humangaussian}.  
Recent methods have also explored learning avatars from large-scale 2D image collections~\cite{hong2023evad, xiu2024puzzleavatar, dong2023ag3d}.

A key limitation in existing works is controllability: prior approaches support conditioning on either text or body shape, but none allow direct, explicit control over detailed clothing items in addition to text and shape. This restricts their application in domains requiring personalized appearance, such as digital fashion or virtual fitting rooms.
We fill this gap by introducing a scalable and fully automatic data generation pipeline that enables the training of generative models conditioned on text, SMPL body shape, and specific clothing images. Our models achieve high-quality 3D human synthesis consistent with all these modalities, offering unprecedented fine-grained control and realism.

\begin{table}[t]
  \centering
    \caption{\textbf{Comparison of related datasets.} Most existing human datasets are limited at scale and none of them provide detailed identity annotation like fine-grained text and clothing image. }
  \resizebox{0.495\textwidth}{!}{
\begin{tabular}{clrcc}
\toprule
\textbf{Type} & \multicolumn{1}{l}{\textbf{Dataset}} & \multicolumn{1}{r}{\textbf{IDs}} & \multicolumn{1}{c}{\textbf{Multi-Text}} & \multicolumn{1}{c}{\textbf{Cloth Assets}} \\
\midrule
\multirow{4}{*}{\rotatebox{90}{3D Scans}} 
      & CustomHuman \cite{ho2023customhuman}  & 80    &  \xmark  & \xmark   \\
      & Sizer~\cite{tiwari2020sizer} & 97 & \xmark   & \xmark \\
      & 2K2K \cite{han20232k2k}  & 2050 &  \xmark    & \xmark  \\
      & THuman2.1 \cite{tao2021thuman2}   & 2500    & \xmark   & \xmark   \\
\midrule
\multirow{7}{*}{\rotatebox{90}{Multi-view Images}} 
      & ActorsHQ~\cite{isik2023actorshq} &  8     & \xmark  &  \xmark   \\
      & ZJU-MoCap~\cite{peng2021neuralbody}    & 10     & \xmark   &   \xmark      \\
      & DNA-Rendering~\cite{chang2023dnarendering}     &  500    & \xmark  & \xmark \\
      & HUMBI~\cite{yu2020humbi}    & 772     & \xmark   & \xmark   \\
      & HuMMan~\cite{cai2022humman}  &  1000    & \xmark    & \xmark \\
      & MVHumanNet~\cite{xiong2024mvhumanet}    &  4500    & \xmark  &   \xmark \\
       & IDOL~\cite{Zhuang2025idol}    & 100K  & \xmark    & \xmark\\
\midrule
\multirow{1}{*}{Ours} 
      & \textbf{InfiniHumanData}  & \textbf{111K} & \cmark  & \cmark \\
    \bottomrule
    \end{tabular}
     }
\label{tab:datasets}
\end{table}

\subsection{Large-Scale 3D Datasets.} 
The availability of high-quality, large-scale 3D datasets is a key driver of progress in generative 3D modeling. While object-centric datasets like Objaverse~\cite{deitke2023objaverse} and ShapeNet~\cite{chang2015shapenet} have enabled remarkable advances for general object synthesis and reconstruction, 3D human datasets pose unique challenges. Commercial human scan repositories such as RenderPeople, Twindom, and Axyz provide highly realistic scans, but are expensive (often $\sim$100 USD per identity). Publicly available 3D human datasets (Tab.~\ref{tab:datasets}) are often constrained by participant recruitment, scanning logistics, and privacy considerations, resulting in limited scale, demographic diversity, and coverage of clothing, age, and body morphology.

Some alternatives use multi-view image capture to reduce costs, but these datasets are typically restricted to fixed camera viewpoints and controlled lighting, limiting their generalizability and value for generative tasks. Recent innovations, such as the IDOL dataset~\cite{Zhuang2025idol}, leverage video diffusion models to synthesize 360-degree images from a single 2D input. However, video diffusion often introduces view inconsistencies and lacks true 3D geometry (see Supp. Mat.), due to neighbor-only attention mechanisms and the absence of explicit 3D supervision.

Critically, existing datasets rarely provide fine-grained annotations of identity level that are essential for training generative models capable of precise control over appearance attributes. Our \textbf{InfiniHumanData} addresses all these limitations by using multi-modal foundation models to generate a large-scale, richly annotated dataset, covering unprecedented diversity across age, ethnicity, body shape, and clothing style, and providing annotations that support high-fidelity, controllable 3D human generation. To accelerate research and enable further expansion, we publicly release our fully automatic data generation pipeline and dataset, empowering the community to create virtually unlimited, realistic human identities.

\section{Method}

\begin{figure*}
    \centering
    \includegraphics[width=\linewidth]{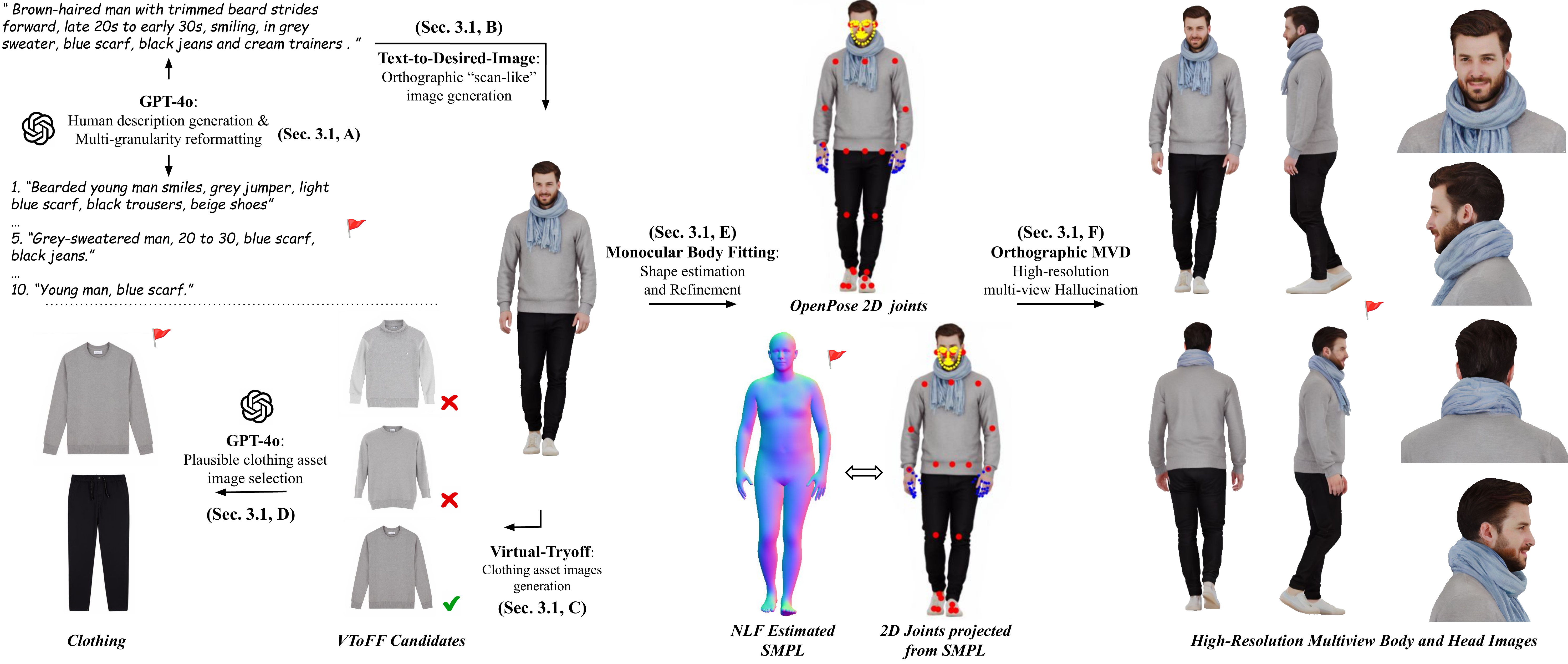}
    \caption{\textbf{Overview of data generation framework} in \textbf{InfiniHumanData}. The process is fully automated by leveraging foundation models. Desired outputs are marked with flags: \textbf{A)} Structured text descriptions, \textbf{C)} Clothing style images, \textbf{E)} Body shape in SMPL format plus face and hand keypoints, \textbf{F)} Orthographic multi-view images with controlled lighting conditions suitable for 3D lifting. }
    \label{fig:infinidata_curation}
\end{figure*}

Our objective is to generate highly realistic 3D avatars that allow precise and flexible control based on multiple user-specified conditions. These conditions include (i) natural language descriptions to define the subject's appearance, (ii) SMPL parameters to govern body shape and pose, and (iii) reference images to specify clothing style. To enable such fine-grained generation, we must model the joint conditional distribution $P({\boldsymbol{y} | \boldsymbol{c}^{\text{text}}, \boldsymbol{c}^{\text{SMPL}}, \boldsymbol{c}^{\text{cloth}}})$, 
where $\mathbf{y}$ represents the generated avatars, and $\mathbf{c}$ terms represent the conditioning signals specified by users.

This task requires a large, diverse, and richly annotated dataset of 3D human avatars, which is costly and impractical to collect and annotate manually. Instead, we present \textbf{InfiniHuman}, a fully automated framework that synthesizes such a dataset by distilling existing foundation models across multiple domains. 
We first detail the construction of our dataset, \textbf{InfiniHumanData}, in Sec.\ref{sec:infinihumandata}, and then describe our controllable generative models, \textbf{InfiniHumanGen}, in Sec.\ref{sec:infinihumangen}.

\subsection{InfiniHumanData - Generation by Reconstruction}
\label{sec:infinihumandata}
To enable highly controllable 3D avatar generation, we first construct a large-scale, richly annotated dataset, \textbf{InfiniHumanData}. Our data generator produces multi-modal outputs for each identity, including structured text descriptions, clothing style images, SMPL body shape and keypoints, and orthographic multi-view images with controlled lighting suitable for 3D lifting (see Fig.~\ref{fig:infinidata_curation} for visualization and detailed breakdown).
In the following, we describe the major components of our data generation pipeline.

\begin{enumerate}[label=\textbf{\Alph*)}, leftmargin=0pt, labelsep=0.5em, itemsep=0.25em, align=left, wide]
\item Multi-Granularity Text Description.
To encode diverse semantic concepts, we design a captioning system that generates both detailed and progressively abstracted descriptions. 
We first caption existing human scan datasets~\cite{tao2021thuman2, ho2023customhuman, han20232k2k} using the protocol from Trellis~\cite{xiang2024trellis}. Next, we randomly sample ten captions and provide them as in-context examples to GPT-4o, prompting it to generate new variations. These generated captions maintain similar lengths and formats, while diversifying attributes such as ethnicity, age group, and clothing style. Each caption is then summarized into ten levels of granularity, ranging from 40 words to 5 words. 
This hierarchical annotation enriches training by exposing models to both coarse (\textit{e.g.}, \textit{old}) and fine-grained (\textit{e.g.}, \textit{late sixties to early seventies}) semantic cues.

\item Orthographic Text-to-Image.
Most text-to-image models (e.g., FLUX) produce images with dramatic perspective and complex lighting, which are suboptimal for 3D reconstruction tasks. To address this, we fine-tune FLUX with a LoRA adapter~\cite{hu2022lora} on orthographic renderings of a few thousand scans under uniform lighting, enabling generation of “scan-like” images (see Fig.~\ref{fig:infinidata_curation}). This stylization step ensures compatibility with downstream 3D lifting processes. In particular, orthographic views are essential for our multi-view diffusion, which relies on simplified epipolar attention~\cite{li2024era3d}. Importantly, this approach preserves the inherent diversity of FLUX while aligning the image domain for reconstruction.
A challenging discriminative user study (Sec.~\ref{sec:evaluation}) further demonstrates that our generated identities achieve visual realism on par with scan renderings.

\item Virtual-TryOff for Clothing Control.
Because a single image can convey garment appearance more precisely than any text description,
we provide direct clothing control by reversing the try-on process. Given a full-body image, we fine-tune OminiControl~\cite{tan2024ominicontrol} to extract a clean garment image via text-based image-to-image translation. This task, termed \textit{Instruct-Virtual-TryOff}, is trained using garment-actor pairs from existing Virtual-TryOn datasets~\cite{choi2021viton, morelli2022dresscode} and prompts like "<Please extract \textit{\{Garment\}} for this person>". 

Each training instance consists of a garment image $I^\text{cloth}$, a corresponding try-on image $I^\text{vton}$, and a textual prompt $e^{\text{text}}$. The model parameters $\boldsymbol{\theta}$ are optimized via the flow-matching objective:
\begin{gather}
\mathcal L_{\text{VToFF}}\!\bigl(\boldsymbol{\theta}\bigr)
  = \mathbb{E}_{t,\boldsymbol{\varepsilon}} \; \!
    \Bigl\lVert
       v_{\boldsymbol{\theta}}\!\bigl(
         \mathbf x_t,\,
         I^{\text{vton}},\,
         e^{\text{text}},\,
         t\bigr)
       -\bigl(\boldsymbol{\varepsilon} - I^{\text{cloth}}\bigr)
    \Bigr\rVert^{2},\\
\text{where}\quad
\mathbf x_t = (1-t)\,I^{\text{cloth}} + t\boldsymbol{\varepsilon}, \; \boldsymbol\varepsilon \sim \mathcal{N}(0,\mathbf I).
\end{gather}
Here, $v_{\boldsymbol{\theta}}$ denotes the network, $\mathbf{x}_t$ is a noisy version of the garment image, and $e^{\text{text}}$ provides the instruction (see Fig.~\ref{fig:infinidata_curation}). The network learns to synthesize clean garment images conditioned on full-body images and textual instructions. This enables users to specify clothing via image, without requiring paired image-scan training data.

\item Negative Samples Rejection.
To remove occasionally wrongly generated images, we use the sampling rejection strategy: first generate four garment images per subject and then employ GPT-4o to select the best match based on considerations like color, texture, length, and detailed features (e.g. zippers, pockets). The detailed prompt for sampling rejection can be found in Supp. Mat.
 
\begin{figure*}
    \centering
    \includegraphics[width=\linewidth]{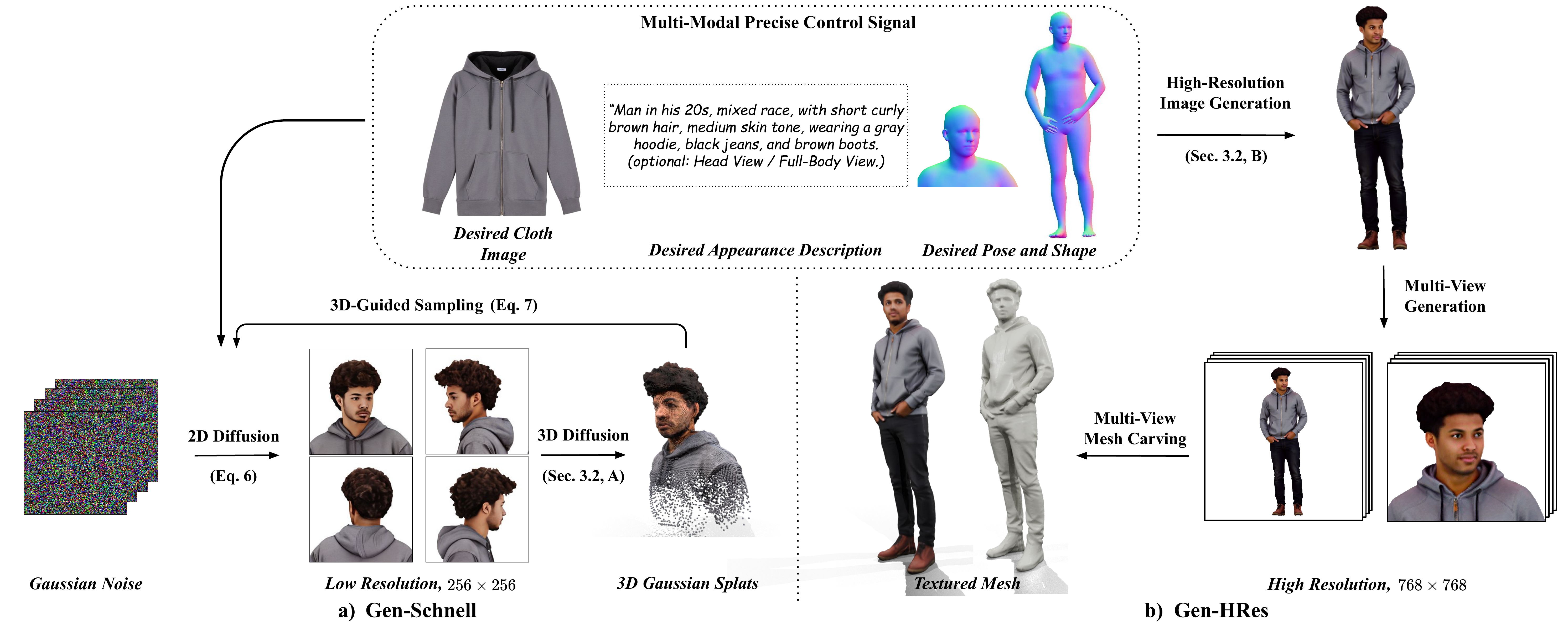}
    \caption{\textbf{Overview of a) Gen-Schnell and b) Gen-HRes in InfiniHumanGen}. Taking text description, explicit SMPL shape, and a cloth image as input, Gen-Schnell generates 3D-GS end-to-end, while Gen-HRes generates high-resolution textured mesh, both matched to input conditions.}
    \label{fig:infinihumangen}
\end{figure*}

\item Monocular Body Fitting for Shape and Pose Control.
We use NLF~\cite{sarandi2024nlf} to regress SMPL parameters from orthographic views by setting FoV to 0.1, followed by refinement via OpenPose 2D joint alignment~\cite{cao2019openpose}. This two-step process ensures that SMPL parameters align accurately with both overall pose and pixel-level features (particularly at face), which is crucial for consistent multi-view generation conditioned on SMPL. 
More specifically, we optimize the SMPL body pose parameters w.r.t. the reprojection error between the orthographically projected SMPL joints and 2D joints estimated by OpenPose:
\begin{equation}
\mathcal L_{\text{reproj}}\!\bigl(\boldsymbol{\theta}\bigr)
  = \sum_{k=1}^{K} w_{ik}
    \Bigl\lVert
      \pi_{\text{ortho}}\!\bigl(J_k(\text{SMPL}\!\left(
         \boldsymbol{\theta_i},
         \boldsymbol{\beta})\right)\bigr)
       -J^\text{OpenPose}_k
    \Bigr\rVert^{2}_{2}
\end{equation}
We carefully tweak the per-joint weights and the regularization to achieve the best pixel-level matching between 3D SMPL and 2D images.
Please refer to Fig.~\ref{fig:ablation_shape} and supplementary material for qualitative visual examples and ablation studies.

\item Orthographic MV-Diffusion.
To produce high-resolution, consistent multi-views, we train a diffusion model on orthographic projections with uniform lighting. Orthographic views have horizontal epipoles, enabling efficient row-wise attention across views.

Given an orthographic RGB image $I^{\text{in}} \in \mathbb{R}^{H \times W \times C}$, our multi-view diffusion (MVD) model generates $N$ views of high-resolution full-body images $I^{\text{body}} \in \mathbb{R}^{N \times H \times W \times C}$ and head images $I^{\text{head}} \in \mathbb{R}^{N \times H \times W \times C}$ from the front, left, right, and back directions. We provide geometric guidance by rendering SMPL normal maps $I^{\text{SMPL}}$ and encoding them, together with the reference image, into the latent space using a pretrained VAE from PSHuman~\cite{li2024pshuman}. For multi-view consistency, we apply orthographic multi-view attention separately within the body and head views, where each row of the each view attends to the same row of other views due to the orthographic constraint across views. Please refer to Fig.~\ref{fig:ablation_multiview} for visual examples. For body-head consistency, we use dense pixel-level cross-attention between corresponding body and head views, where each pixel of body image attends to pixels of the head image under the same view. 
The UNet denoiser $\epsilon(\boldsymbol \theta)$ is fine-tuned using the following objective:
\begin{gather}
\mathcal{L}_{\text{MVD}}\!\bigl(\boldsymbol{\theta}\bigr)
  = \mathbb{E}_{t,\boldsymbol{\varepsilon}} 
  \sum_{p \in \{\text{body}, \text{head}\}}
    \Bigl\lVert
       \epsilon_{\boldsymbol{\theta}}\!\bigl(
         \mathbf{x}_t^p,\,
         I^{\text{in}},\,
         I^{\text{SMPL}},\,
         t\bigr)
       -\boldsymbol{\varepsilon}
    \Bigr\rVert^{2},\\
\text{where}\quad
\mathbf{x}_t^p = \sqrt{\bar{\alpha}_t} I^p + \sqrt{1-\bar{\alpha}_t} \boldsymbol\varepsilon, \quad \boldsymbol\varepsilon \sim \mathcal{N}(0,\mathbf{I}).
\end{gather}
Here, $\bar{\alpha}_t$ determines the noise level at each diffusion step $t$. At inference time, our multi-view diffusion model takes an orthographic input image and SMPL normal maps as input, generating high-resolution multi-view body and head images (see Fig.~\ref{fig:infinidata_curation}, right).
\end{enumerate}

\subsection{InfiniHumanGen - Generation with Precise Control}
\label{sec:infinihumangen}

\subsubsection{Joint Conditional Distribution.}
Leveraging InfiniHumanData, which contains 111K diverse identities each annotated with multi-granularity text captions $\boldsymbol{c}^{\text{text}}$, SMPL parameters $\boldsymbol{c}^{\text{SMPL}}$, corresponding cloth images $\boldsymbol{c}^{\text{cloth}}$, and orthographic multi-view images $\boldsymbol{y}^{{mv}}$, we learn a joint conditional distribution $P({\boldsymbol{y} | \boldsymbol{c}^{\text{text}}, \boldsymbol{c}^{\text{SMPL}}, \boldsymbol{c}^{\text{cloth}}})$ to enable precise avatar generation. 
We train two complementary models to support both fast and high-fidelity generation:

\begin{enumerate}[label=\textbf{\Alph*)}, leftmargin=0pt, labelsep=0.5em, itemsep=0.25em, align=left, wide]
\item Gen-Schnell: Fast End-to-End Generation.
Gen-Schnell is a low-latency model that directly generates 3D avatars as Gaussian splats~\cite{Kerbl20233dgs}. Inspired by Human-3Diffusion~\cite{xue2024human3diffusion}, we combine 2D multi-view generation (from MVDream~\cite{shi2024mvdream}) with a splatting decoder that enforces consistency across views.
To inject condition signals, we encode SMPL normal maps $\boldsymbol{c}^{\text{SMPL}}$ and clothing images $\boldsymbol{c}^{\text{cloth}}$ using the MVDream VAE, and concatenate them channel-wise with the initial noise $\mathbf{x}_t$. The 2D diffusion model $\epsilon(\boldsymbol{\theta})$ predicts noise values, which are used to reconstruct clean multi-view images $\tilde{\mathbf{x}}_0$:
\begin{equation}
    \tilde{\mathbf{x}}_0=\frac{1}{\sqrt{\bar{\alpha}_t}}\left(\mathbf{x}_t-\sqrt{1-\bar{\alpha}_t} {\epsilon}_{\boldsymbol\theta} \left(\mathbf{x}_t, \mathbf{c}^{\text{text}},\mathbf{c}^{\text{SMPL}}, \mathbf{c}^{\text{cloth}} , t\right)\right).
\end{equation}
While the resulting multi-view images $\tilde{\mathbf{x}}_0$ provide strong shape priors, they may exhibit inconsistencies across views. To address this, our 3D-GS generator $g(\boldsymbol\phi)$ takes the predicted multi-view images $\tilde{\mathbf{x}}_0$ and initial noise $\mathbf{x}_t$ to generate consistent 3D Gaussian splats $\hat{\mathcal{G}}_0$, which render consistent multi-view images $\hat{\mathbf{x}}_0$. During each sampling step from $t$ to $t-1$, we replace 2D predictions with 3D-GS rendered images to ensure consistency:
\begin{equation}
\begin{aligned}
\mu_{t-1}\!\left(\mathbf{x}_t, \hat{\mathbf{x}}_0\right)
  &= \frac{\sqrt{\alpha_t}\bigl(1-\bar{\alpha}_{t-1}\bigr)}{1-\bar{\alpha}_t}\,
     \mathbf{x}_t
   + \frac{\sqrt{\bar{\alpha}_{t-1}}\,\beta_t}{1-\bar{\alpha}_t}\,
     \hat{\mathbf{x}}_0, \\
\mathbf{x}_{t-1}
  &\sim
  \mathcal{N}\!\bigl(
      \mathbf{x}_{t-1};
      \tilde{\boldsymbol{\mu}}_t\!\left(\mathbf{x}_t, \hat{\mathbf{x}}_0\right),
      \tilde{\beta}_{t-1}\mathbf{I}
  \bigr).
\end{aligned}
\end{equation}
At the final timestep $t=0$, $\hat{\mathcal{G}}_0$ is output as the final 3D-GS, see Fig.~\ref{fig:infinihumangen}. Gen-Schnell is highly efficient and produces 3D-GS in about 12 seconds. However, due to the low-resolution constraint of MVDream (256×256), detailed features (e.g., facial textures, textual elements) appear blurry, motivating our high-resolution generator, Gen-HRes.

\begin{figure*}
    \centering
    \includegraphics[width=\linewidth]{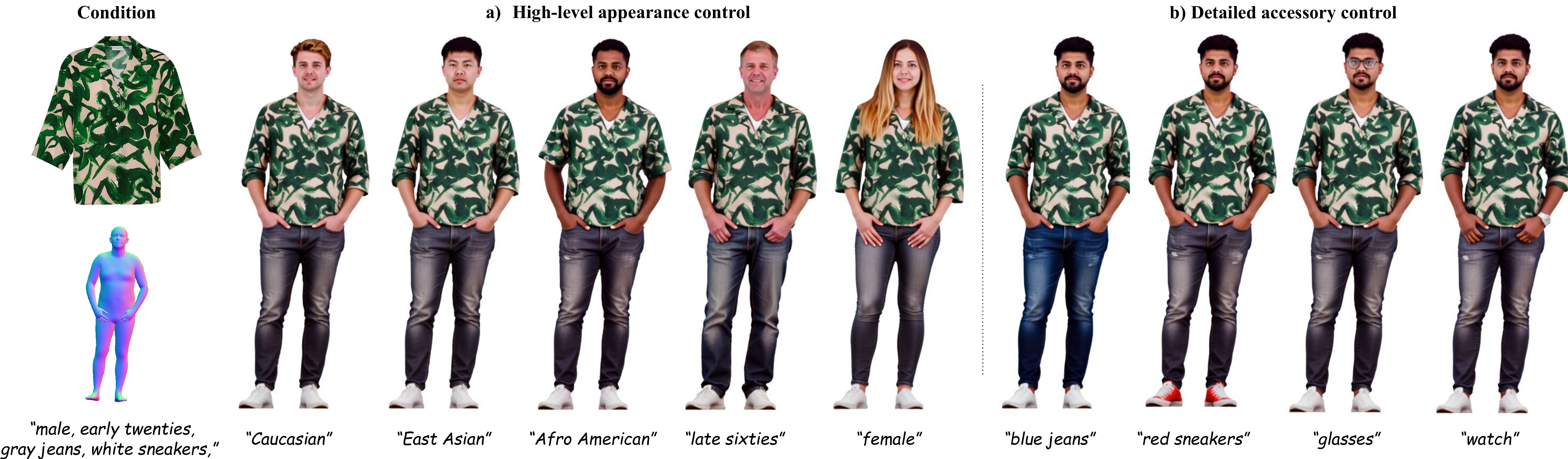}
    \caption{\textbf{Fine-grained text controllability in Gen-HRes} over (a) overall subject identity, such as ethnicity, age, gender, etc. By fixing the initial Gaussian noise, Gen-HRes can generate (b) same identity with different detailed accessory appearance, such as watch, glasses, and colors of wearing assets.}
    \label{fig:text_control}
\end{figure*}

\item Gen-HRes: High-Resolution Generation.
For photorealistic avatar generation from multiple conditions, Gen-HRes frames it as a multi-image-to-image translation task, where we fine-tune OminiControl2~\cite{tan2025ominicontrol2} on InfiniHumanData. Using full-body images $\mathbf{y}^{\text{2D}}$ as target, we optimize the flow matching objective:
\begin{gather}
\mathcal L_{\text{HRes}}\!\bigl(\boldsymbol{\theta}\bigr)
  = \mathbb{E}_{t,\boldsymbol{\varepsilon}} \; \!
    \Bigl\lVert
       v_{\boldsymbol{\theta}}\!\bigl(
         \mathbf x_t,\,
         \mathbf{c}^{\text{text}},\,
         \mathbf{c}^{\text{cloth}},\,
         \mathbf{c}^{\text{SMPL}},\,
         t\bigr)
       -\bigl(\boldsymbol{\varepsilon} - \mathbf{y}^{\text{2D}}\bigr)
    \Bigr\rVert^{2},\\
\text{where}\quad
\mathbf x_t = (1-t)\,\mathbf{y}^{\text{2D}} + t\boldsymbol{\varepsilon}, \; \boldsymbol\varepsilon \sim \mathcal{N}(0,\mathbf I).
\end{gather}
Our data and model design ensure that the generated multi-view images are well aligned with the conditioning SMPL mesh. This alignment allows us to compute surface normals with Sapiens2B~\cite{khirodkar2024sapiens} and apply SMPL-driven volumetric carving via PSHuman~\cite{li2024pshuman} for high-fidelity 3D mesh reconstruction.

Compared to Gen-Schnell, Gen-HRes not only achieves higher resolution and visual fidelity, but also supports detailed text prompting. By fixing initial Gaussian noise, Gen-HRes can precisely control fine-grained attributes, such as glasses or garment colors, through detailed text descriptions, as shown in Fig.~\ref{fig:text_control}.
Gen-HRes enables high-fidelity avatar generation in approximately 4 minutes.
\end{enumerate}

\section{Experiments}

\subsection{Implementation Details}
The orthographic multiview diffusion model used in InfiniHuman and Gen-HRes is built upon the pre-trained text-to-image model SD2.1-unclip. We concatenate the input image latents with noise latents along the channel dimension. The noise latents are replicated for each view, and the text embedding is repurposed to generate distinct head and body views, similar to PSHuman~\cite{li2024pshuman}. Our model generates four orthogonal body views and four head views from single orthographic input body image.
For the orthographic multiview diffusion models, we train on 8 H100 with effective batch size of 128 for 2 days on orthographic uniform lighting rendering of 6000 high-quality human scans from Twindom, CustomHuman, and THuman2.1~\cite{tao2021thuman2, ho2023customhuman, twindom}. 
We use front-view renders with text labels to fine-tune Flux-Dev~\cite{flux2024} LoRA for the orthographic text-to-image task.
For constructing the InfiniHumanData, we use GPT-4o to enhance correctness of automatic cloth labeling, where each subject takes around \$0.03.
Based on InfiniHumanData, we train Gen-Schnell on 8 A100 GPUs with effective batch size 256 over approximately 2 days, and Gen-HRes on 2 H100 GPUs with effective batch size of 32 for 2 days. Please refer to Supp. Mat. for implementation details on Gen-Schnell as well as Gen-HRes, and for prompting details on constructing InfiniHumanData.

\subsection{Evaluation Benchmark}
\label{sec:evaluation}

\begin{table}[t]
  \caption{\textbf{Quantitative comparison results.} We report user study results for appearance quality and text alignment, where most participants prefer our method. We also achieve SOTA in T2I metrics such as CLIP and FID.}
\centering
  \resizebox{0.96\columnwidth}{!}{%
  \begin{tabular}{@{}lccccc c@{}}
    \toprule
    \multirow{2}{*}{Method} &
      \multicolumn{1}{c}{Quality$\uparrow$} &
      \multicolumn{1}{c}{Alignment$\uparrow$} &
      \multirow{2}{*}{FID$\downarrow$} &
      \multirow{2}{*}{CLIP Score$\uparrow$} &
      \multirow{2}{*}{Runtime$\downarrow$} \\
    & (User Study) & (User Study) & & & \\ \midrule
    MVDream               & \cellcolor{heatOrange}20.83\% & \cellcolor{heatOrange}20.36\% & \cellcolor{heatOrange}141.33 & \cellcolor{heatOrange}30.37 & \cellcolor{heatRed}\textbf{2.8s}\\
    SPAD                  & 2.02\% & 1.55\% & 150.43 & 28.58 & 13.9s \\
    \textbf{\textit{Gen-Schnell}} & \cellcolor{heatRed}\textbf{77.14}\% & \cellcolor{heatRed}\textbf{78.10}\% & \cellcolor{heatRed}\textbf{100.39} & \cellcolor{heatRed}\textbf{30.82} & \cellcolor{heatOrange}12.9s \\ \midrule
    TADA                  & 1.27\% & 1.27\% & \cellcolor{heatYellow}129.68 & \cellcolor{heatYellow}28.84 &  213m \\
    DreamAvatar           & 1.27\% & \cellcolor{heatYellow} 1.90\% & 151.57 & 28.42 & 384m \\
    HumanGaussian         & \cellcolor{heatYellow}2.22\% & \cellcolor{heatOrange}3.48\% & 140.24 & \cellcolor{heatRed}\textbf{30.56} & \cellcolor{heatOrange} 40 m \\
    HumanNorm             & \cellcolor{heatOrange}2.54\% & \cellcolor{heatOrange}3.48\% & \cellcolor{heatOrange}101.84 & 28.30 & 117 m \\
    AvatarVerse           & 0.32\% & 0.32\% & 156.52 & 28.69 & \cellcolor{heatYellow} 44 m \\
    \textbf{\textit{Gen-HRes}}     & \cellcolor{heatRed}\textbf{92.39}\% & \cellcolor{heatRed}\textbf{89.56}\% & \cellcolor{heatRed}\textbf{82.28} & \cellcolor{heatOrange}30.43 & \cellcolor{heatRed}\textbf{4 m} \\
    \bottomrule
  \end{tabular}}
  \label{tab:comparison}
\end{table}

\begin{figure*}
    \centering
    \includegraphics[width=\linewidth]{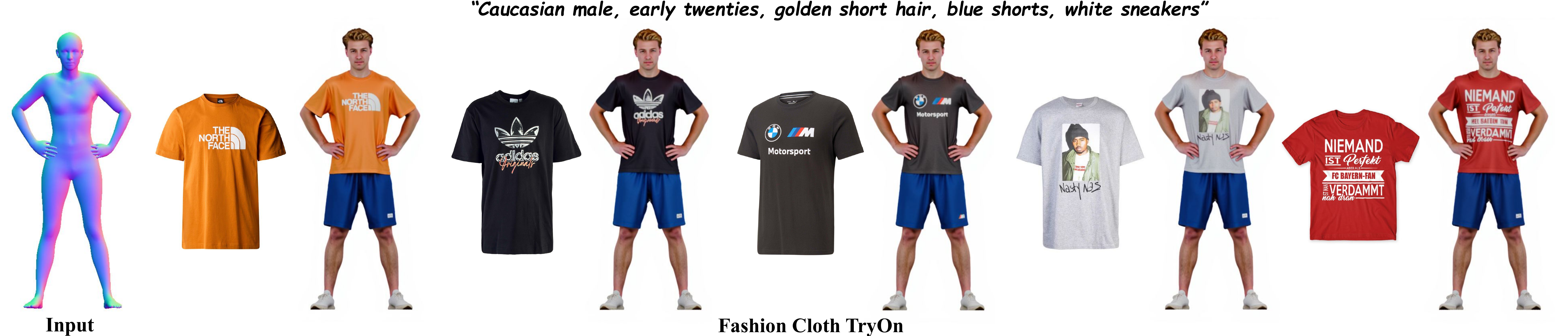}
    \caption{\textbf{Generate avatars with given garment from fashion industry}. The identity is preserved while TryOn garment is changing.}
    \label{fig:demo_tryon}
\end{figure*}

\begin{figure*}
    \centering
    \includegraphics[width=\linewidth]{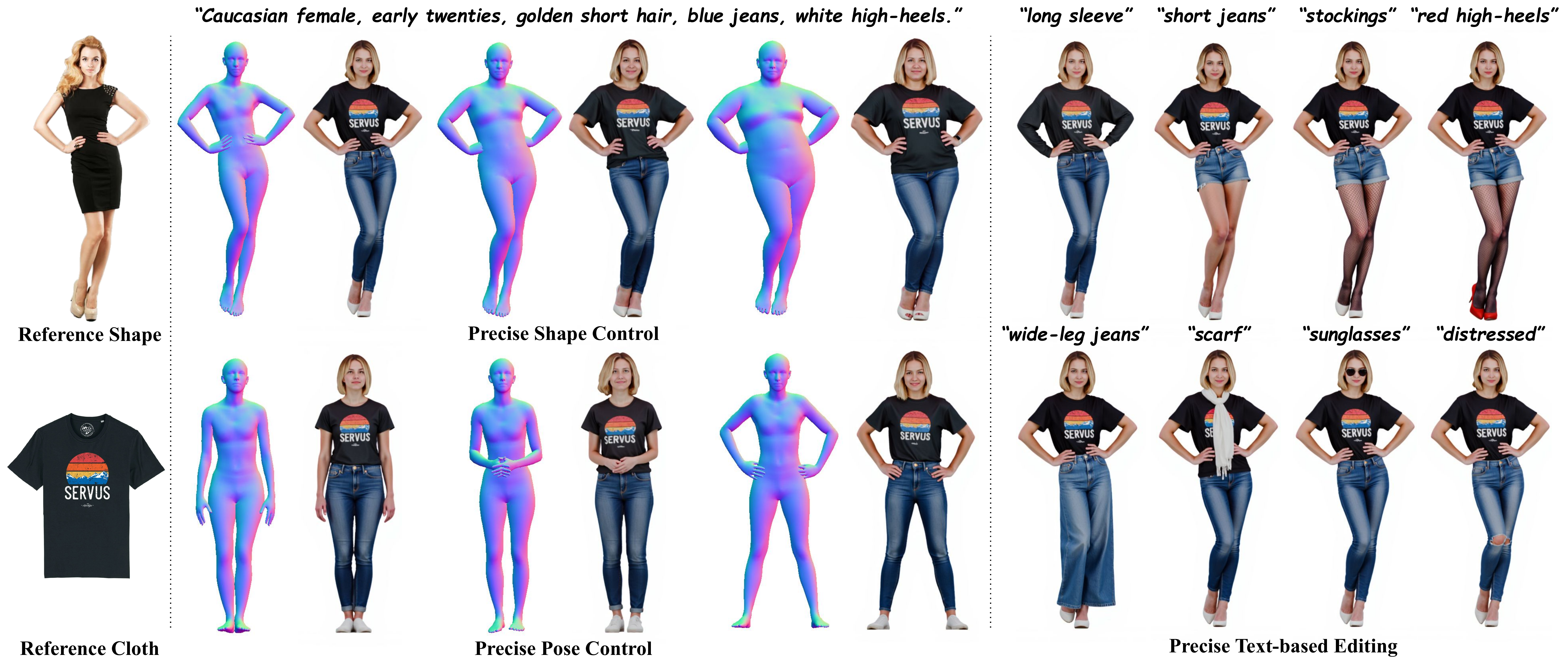}
    \caption{\textbf{Generate avatars with precise pose shape control and text-based editing}. The identity is preserved during shape and text-based editing.}
    \label{fig:demo_shape}
\end{figure*}
We compare Gen-Schnell with feed-forward text-based multi-view generation approaches such as MVDream~\cite{shi2024mvdream} and SPAD~\cite{kant2024spad}, which can generate multi-view images from text prompt within a minute. We compare Gen-HRes with SDS-based text-to-avatar approaches, e.g. DreamAvatar~\cite{cao2024dreamavatar}, AvatarVerse~\cite{zhang2024avatarverse}, HumanGaussian~\cite{liu2024humangaussian}, HumanNorm~\cite{huang2024humannorm}, and TADA~\cite{liao2024tada}. These optimization-based methods achieve higher quality than feed-forward approaches but typically require several hours for generation. Therefore, we also compare with Chupa~\cite{kim2023chupa}, a mesh-based avatar generator directly learned from 3D scans. Furthermore, we evaluate the realism of generated identities in InfiniHumanData.

\subsubsection{Qualitative Comparison.} As depicted in Fig.~\ref{fig:comparison}, Gen-HRes has various advantages over baselines: (1) multi-view consistency, because Gen-HRes generates textured mesh as output, while SDS-method optimizes per view given text prompt, which can lead to the Janus problem. (2) enhanced realism, Gen-HRes does not suffer from unnatural saturation, which is a typical problem in SDS-based generation. (3) text-following ability, Gen-HRes leverages foundational text-based generation capability from FLUX, which shows stronger text-following ability than SDS-based methods, especially in details such as color of garments. Gen-Schnell also shows better text-following ability (e.g. head view, color) than previous works. Please refer to Fig.3 and Fig.4 in Supp. Mat. for more comparison.

\subsubsection{Quantitative Comparison.} We conducted a user study to quantitatively compare with SOTA methods in text-based generation. We asked 42 participants to evaluate videos rendered from generated 3D avatars generated by different methods and to vote for the best methods based on overall appearance quality and alignment with text description. We also report quantitative numbers in FID between generated results and rendered images from human scans. Additionally, we use the CLIP Score to quantify the semantic alignment between the text description and the renderings.
Tab.~\ref{tab:comparison} presents average scores across 32 prompts. The results of user studies, FID, and CLIP demonstrate that our Gen-Schnell and Gen-HRes outperforms SOTA feed-forward text-to-3D and SDS-based text-to-3D avatar methods, respectively. Our method achieves the highest overall result quality and the most accurate alignment with the prompt's semantics. More importantly, our Gen-HRes can generate 3D avatars with at least 8 times less computational time than high-resolution baselines, demonstrating that our InfiniHumanGen is the most efficient high-resolution avatar generative model.

\subsubsection{InfiniHumanData Evaluation.} To assess the realism of InfiniHumanData, we conducted a user study comparing it against renderings from real human scans. The goal was to evaluate whether users could distinguish our generated avatars from those based on actual 3D scan data.
Participants were presented with image pairs: one image rendered from a real scan, and the other randomly sampled from InfiniHumanData. For each pair, users were asked to select the more realistic image, or choose “both” if they could not tell the difference.
Across all trials, real scan renderings received \textbf{\textit{746}} votes, while InfiniHumanData images received \textbf{\textit{765}} votes. The small difference in votes indicates that our InfiniHumanData achieves a high degree of visual realism, closely matching the appearance of scans.

\subsection{Fine-grained Controllability}
\subsubsection{Precise Clothing Control.} As shown in Fig.~\ref{fig:demo_tryon}, Gen-HRes can generate avatars with high fidelity to the input clothing images. By fixing the initial Gaussian noise, we can generate the same subject wearing different garments, preserving identity across try-on results. This demonstrates strong, identity-preserving clothing controllability.

\subsubsection{Precise Pose and Shape Control.} As illustrated in Fig.~\ref{fig:demo_shape}, Gen-HRes accurately follows the body shape and pose specified by the SMPL condition, faithfully transferring to the generated avatar.

\subsubsection{Precise Text-based Generation and Editing.} Gen-HRes enables control over high-level human attributes, such as ethnicity, age, and gender, all through text input (see Fig.~\ref{fig:text_control}). More importantly, it supports fine-grained text-based editing while maintaining identity consistency. As shown in Fig.~\ref{fig:demo_shape}, we generate the same subject with different accessories, such as stockings, scarves, or sunglasses.

\subsection{Application}
\subsubsection{TryOn from Photographs.} Our \textit{Instruct-Virtual-TryOff} module demonstrates strong generalization: it can extract clean garment images directly from real-world photographs. As shown in Fig.~\ref{fig:demo_real}, we extract clothing assets from photo captures and generate corresponding avatars with user-specified text controls.

\subsubsection{Re-animation.} Leveraging the underlying SMPL parametric body, our generated 3D avatars can be reanimated using SMPL motion data by barycentric interpolation of SMPL skinning weights onto the generated mesh surface. See Fig.~\ref{fig:animation_fabrication} for re-animation examples.

\subsubsection{Figurine Fabrication.} Gen-HRes produces high-quality, watertight 3D meshes, enabling direct 3D printing of physical figurines. The printed figurines are physically robust and can stand independently, as shown in Fig.~\ref{fig:animation_fabrication}, demonstrating the real-world physical compatibility~\cite{[guo2024physical3d]} of generated avatars.

\subsection{Ablation Study}

We qualitatively ablate different design choices, showcasing the importance of orthographic MVD (Sec.~\ref{sec:infinihumandata}F, Fig.~\ref{fig:ablation_multiview}), generating
scan-like images (Sec.~\ref{sec:infinihumandata}B, Figs.~\ref{fig:ablation_multiview},~\ref{fig:ablation_desired_imgs}), additional SMPL fitting
(Sec.~\ref{sec:infinihumandata}E, Fig.~\ref{fig:ablation_shape}), and tolerance to inaccurate SMPL for children
generation (Fig.~\ref{fig:ablation_shape}). Please refer to individual figures for examples.

\section{Limitations and Future Works}

Although our Gen-HRes can perform high-fidelity generation, it is still slower than the end-to-end 3D generation pipeline, Gen-Schnell. However, Gen-Schnell cannot generate faithful details such as face because of the low-resolution (256$\times$256) of pretrained MVDream. 
Due to limited training resources, we cannot directly train a higher-resolution Gen-Schnell. However, we publicly release all high-resolution (768$\times$768) InfiniHumanData with multi-modal labels.
Future works can consider training a high-resolution text-based 3D-GS model, which achieves fast and high-quality end-to-end multi-modal avatar generation.

\begin{figure*}[t]
  \centering

  \noindent\begin{minipage}{\textwidth}
    \centering
    \includegraphics[width=0.9\linewidth]{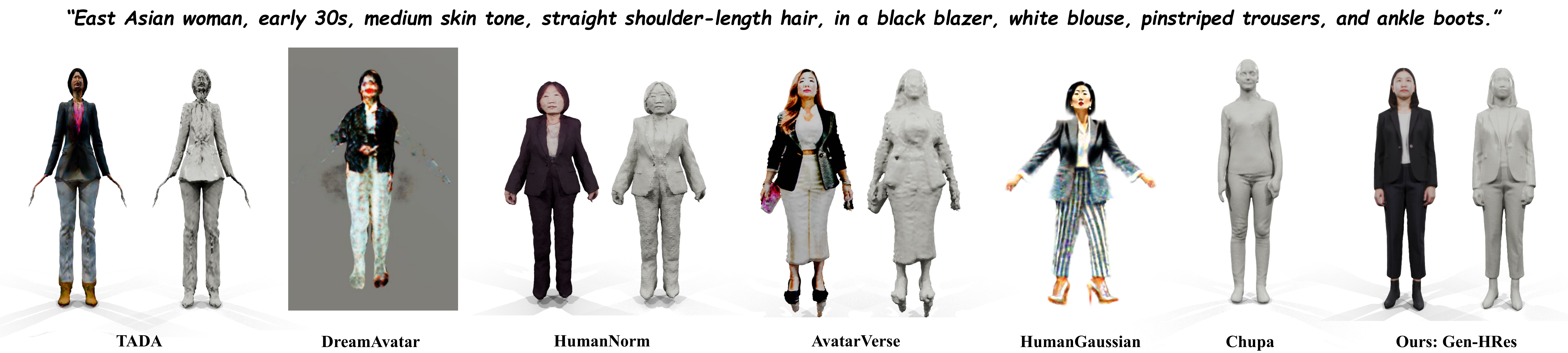}
    \caption{\textbf{Qualitative comparison to SOTA text-to-3D avatar generators.} We compare with SDS-based avatar generation methods and a mesh-based avatar generation method Chupa~\cite{kim2023chupa}. Our generator can follow the text very well and also achieve outstanding generation quality.}
    \label{fig:comparison_genhres_chupa}
  \end{minipage}

  \noindent\begin{minipage}{\textwidth}
    \centering
    \includegraphics[width=0.9\linewidth]{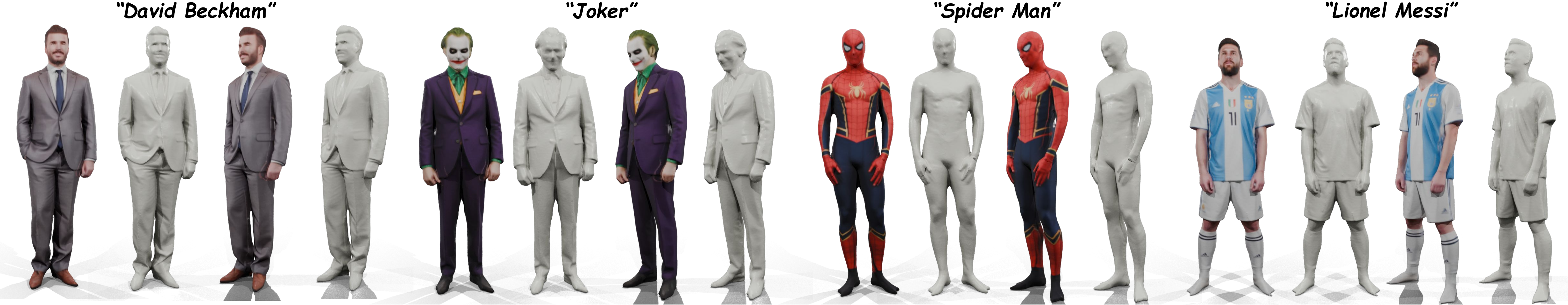}
     \caption{\textbf{Generated famous people and characters by names in Gen-HRes}. Please zoom in for details.}
    \label{fig:demo_famous}
  \end{minipage}

  \noindent\begin{minipage}{\textwidth}
    \centering
    \includegraphics[width=0.9\linewidth]{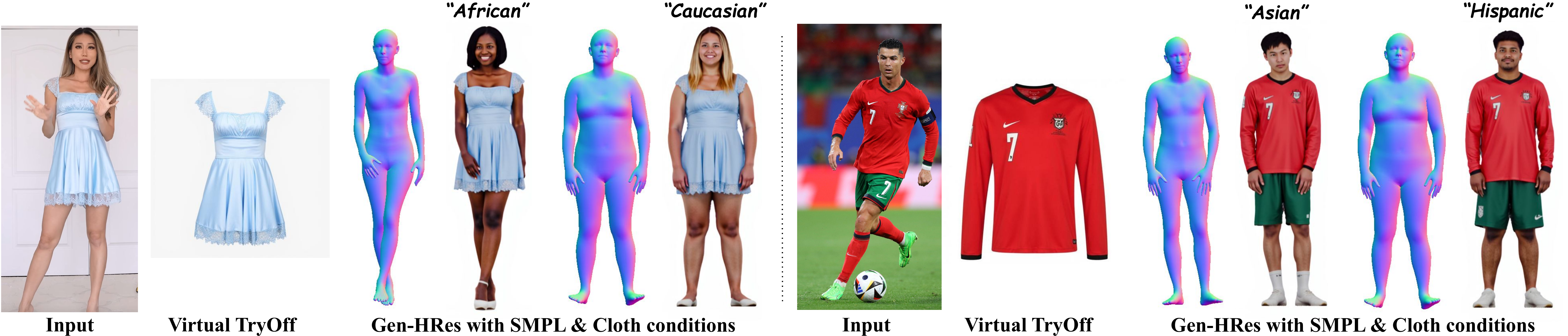}
    \caption{\textbf{Generated avatars with garments in real person photos}. We can extract clean garments from photos and use for TryOn. \href{https://www.shutterstock.com/image-photo/leipzig-germany-june-18-2024-cristiano-2480454921}{Images from Shutterstock}.}
    \label{fig:demo_real}
  \end{minipage}

  \noindent\begin{minipage}{\textwidth}
    \centering
    \includegraphics[width=0.9\linewidth]{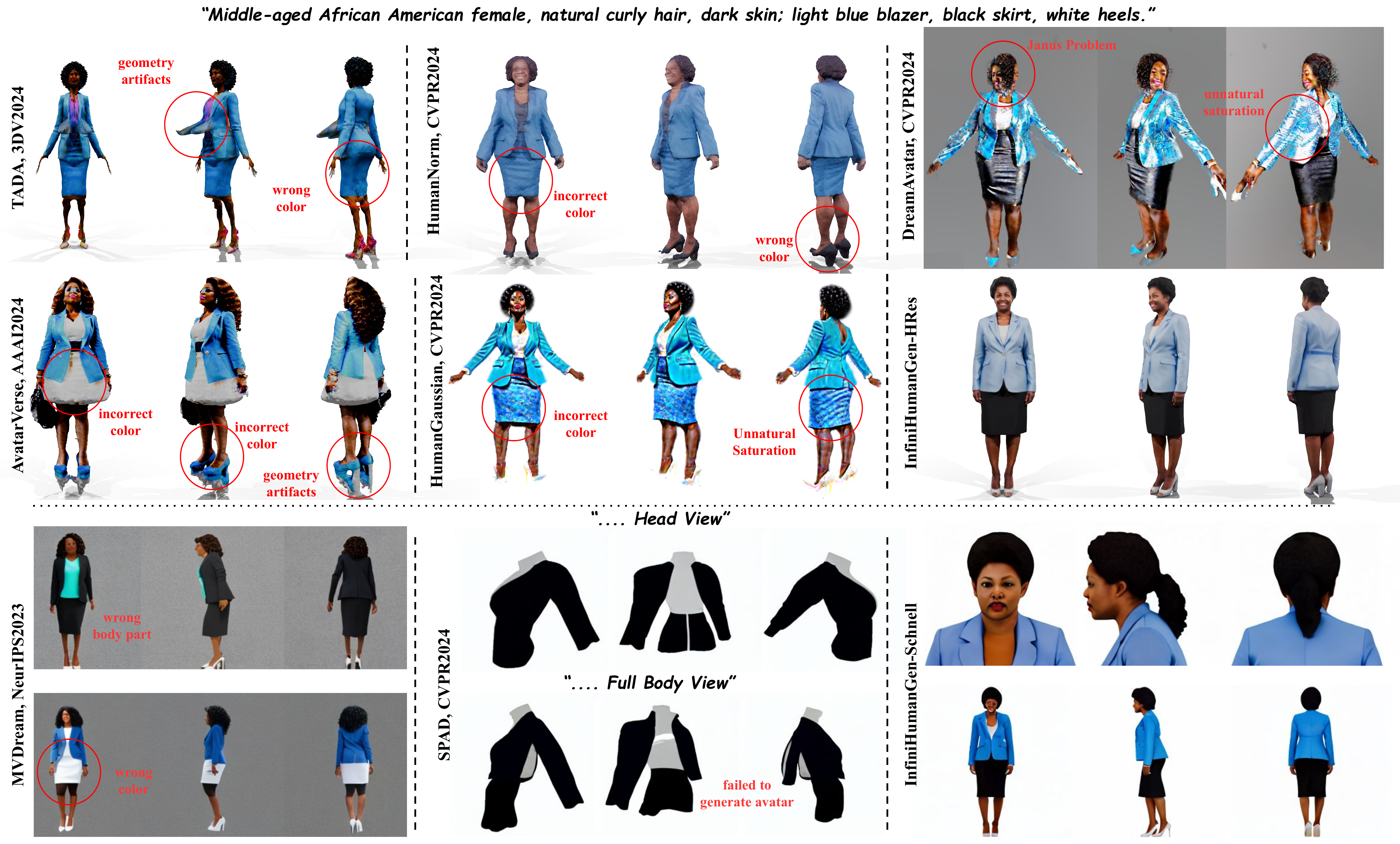}
    \captionof{figure}{\textbf{Qualitative comparison to SOTA text-to-3D avatar approaches.}
      Gen-HRes avoids Janus/artifacts, aligns to prompts, and is $8\times$ faster (Tab.~\ref{tab:comparison}).}
    \label{fig:comparison}
  \end{minipage}

\end{figure*}

As shown in Fig.~\ref{fig:demo_famous}, our pipeline can generate famous people by names. However, GPT-4o refuses to identify unmatched samples because of privacy issues. Future works may adopt a different vision-language model to include famous names in InfiniHumanData. Moreover, our Gen-HRes adopts multi-view mesh carving to obtain textured mesh from orthographic views, which can cause texture artifacts in self-occluded parts of the avatar. Future works may consider a data-driven approach for the mesh reconstruction from multi-view images.

\section{Conclusion}
In this work, we present InfiniHuman, a novel framework for realistic and highly controllable 3D avatar generation. To overcome the fundamental challenge of scarce and expensive annotated human data, we developed a fully automated data generation framework that repurposes multiple pretrained foundation models. This enables the creation of InfiniHumanData, a large-scale, richly annotated dataset with 111K diverse identities and comprehensive control signals.
Building on this foundation, our InfiniHumanGen framework delivers rapid, high-fidelity avatar synthesis with unprecedented fine-grained control, enabling users to specify appearance, shape, pose, and clothing through intuitive multi-modal inputs. Extensive experiments demonstrate that InfiniHuman not only outperforms prior methods in visual quality and speed, but also sets a new standard for precise, attribute-level controllability in 3D human generation.
Importantly, our approach democratizes high-quality avatar creation via an accessible and scalable solution. To support further research and broad adoption, we will publicly release InfiniHumanData, InfiniHumanGen, and our automatic data generation pipeline, empowering the community to create unlimited, realistic, and diverse 3D humans with full user control.

\begin{acks} 
This work is made possible by funding from the Carl Zeiss Foundation. 
This work is also funded by the Deutsche Forschungsgemeinschaft (DFG, German Research Foundation) - 409792180 (EmmyNoether Programme, project: Real Virtual Humans) and the German Federal Ministry of Education and Research (BMBF): Tübingen AI Center, FKZ: 01IS18039A. 
The authors thank the International Max Planck Research School for Intelligent Systems (IMPRS-IS) for supporting Y.Xue.
G. Pons-Moll is a member of the Machine Learning Cluster of Excellence, EXC number 2064/1 – Project number 390727645. 

Y.Xue is the first author and corresponding author. Y.Xue initialized the core idea, organized the project, developed the current method, conducted experiments, and wrote the draft.
X.Xie contributed to the method development and draft writing. M.Kostyrko contributed to the teaser figure rendering and re-animation of Gen-HRes avatars. All the team members made necessary contributions to the method development and paper writing.
\end{acks}

\newpage

\captionsetup[figure]{skip=2pt}
\captionsetup[subfigure]{skip=1pt}

\newpage

\begin{figure*}
    \centering
    \includegraphics[width=\linewidth]{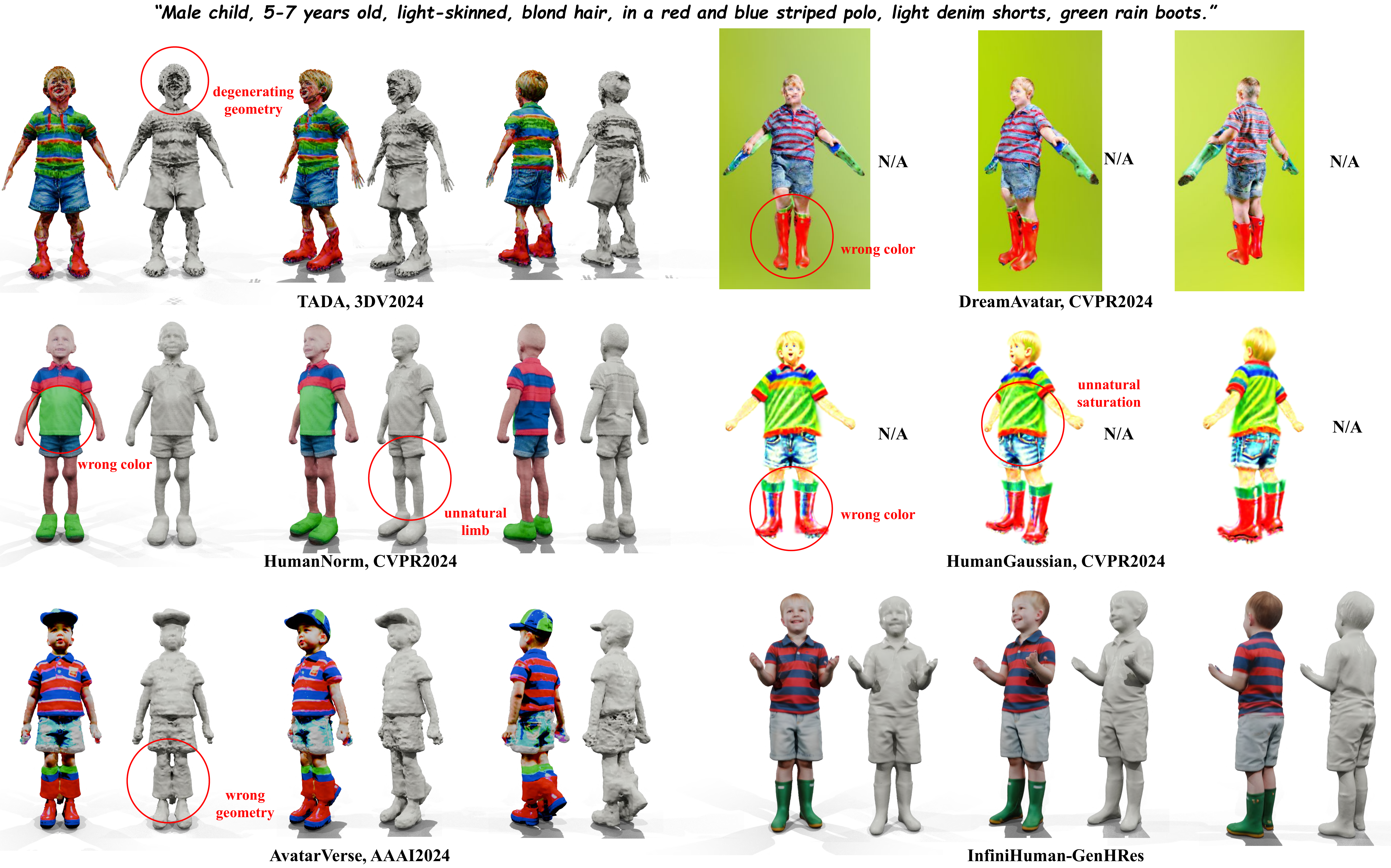}
    \caption{\textbf{Qualitative appearance and geometry comparison to SOTA text-to-3D avatar approaches}. Please refer to Supp. Mat. for more comparisons.}
    \label{fig:comparison_geometry}
\end{figure*}

\begin{figure*}                %
  \centering

  \begin{minipage}{0.48\linewidth}
    \centering
    \includegraphics[width=\linewidth]{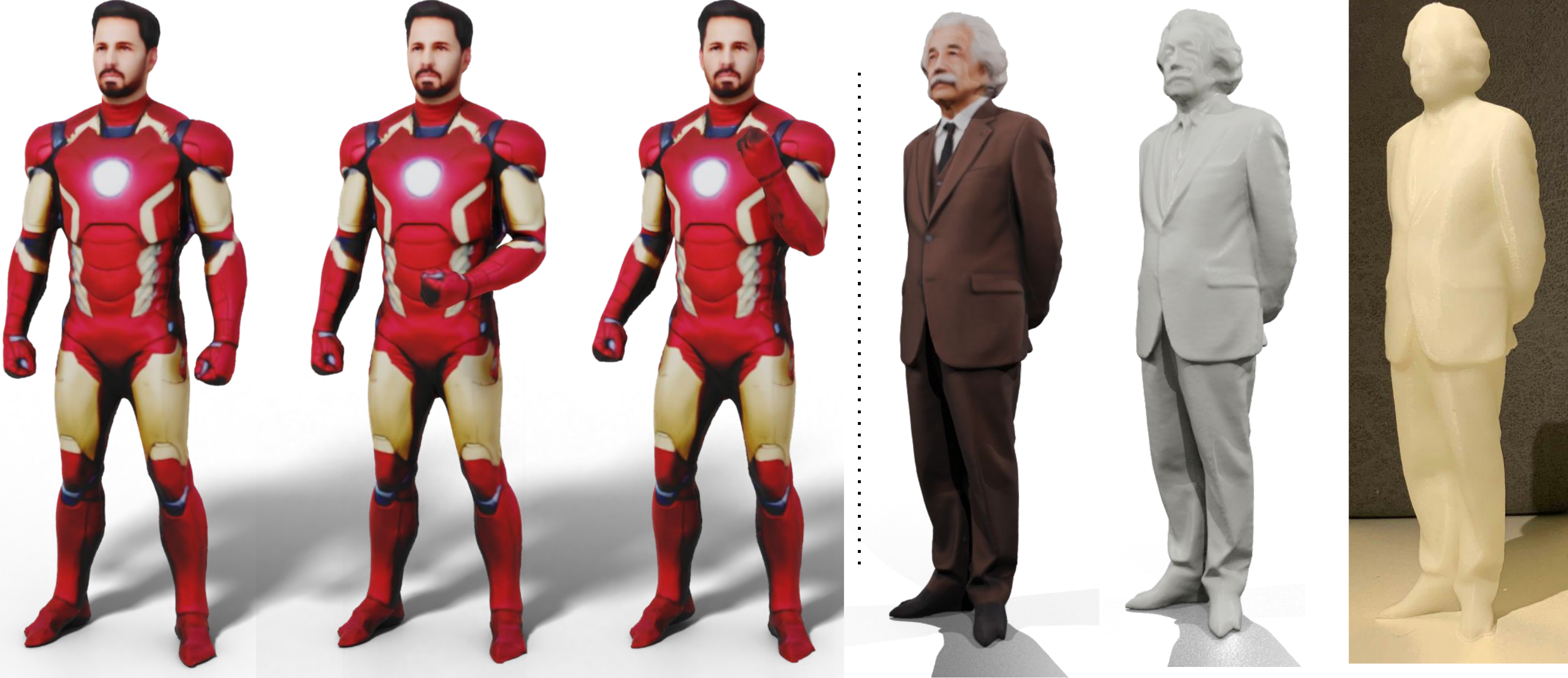}
    \captionof{figure}{Re-animation (left) and Fabrication (right) of Gen-HRes avatars.}
    \label{fig:animation_fabrication}
  \end{minipage}%
  \hfill
  \begin{minipage}{0.48\linewidth}
    \centering
    \includegraphics[width=\linewidth]{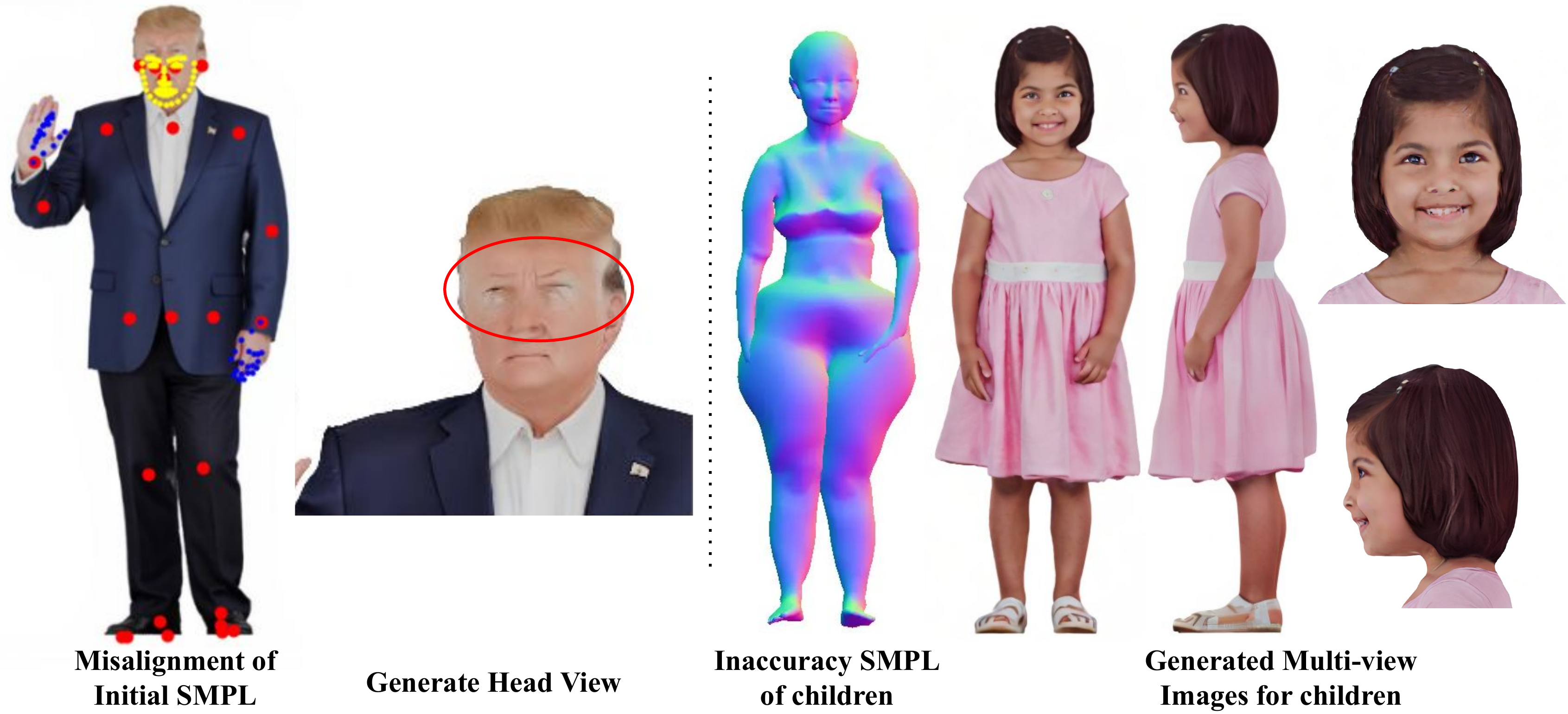}
    \captionof{figure}{Misaligned joints cause bad face generation (left). Our pipeline tolerates bad SMPL estimation for children, yielding good multi-views (right).}
    \label{fig:ablation_shape}
  \end{minipage}

\end{figure*}

\begin{figure*}                %
  \centering

  \begin{minipage}{0.46\linewidth}
    \centering
    \includegraphics[width=\linewidth]{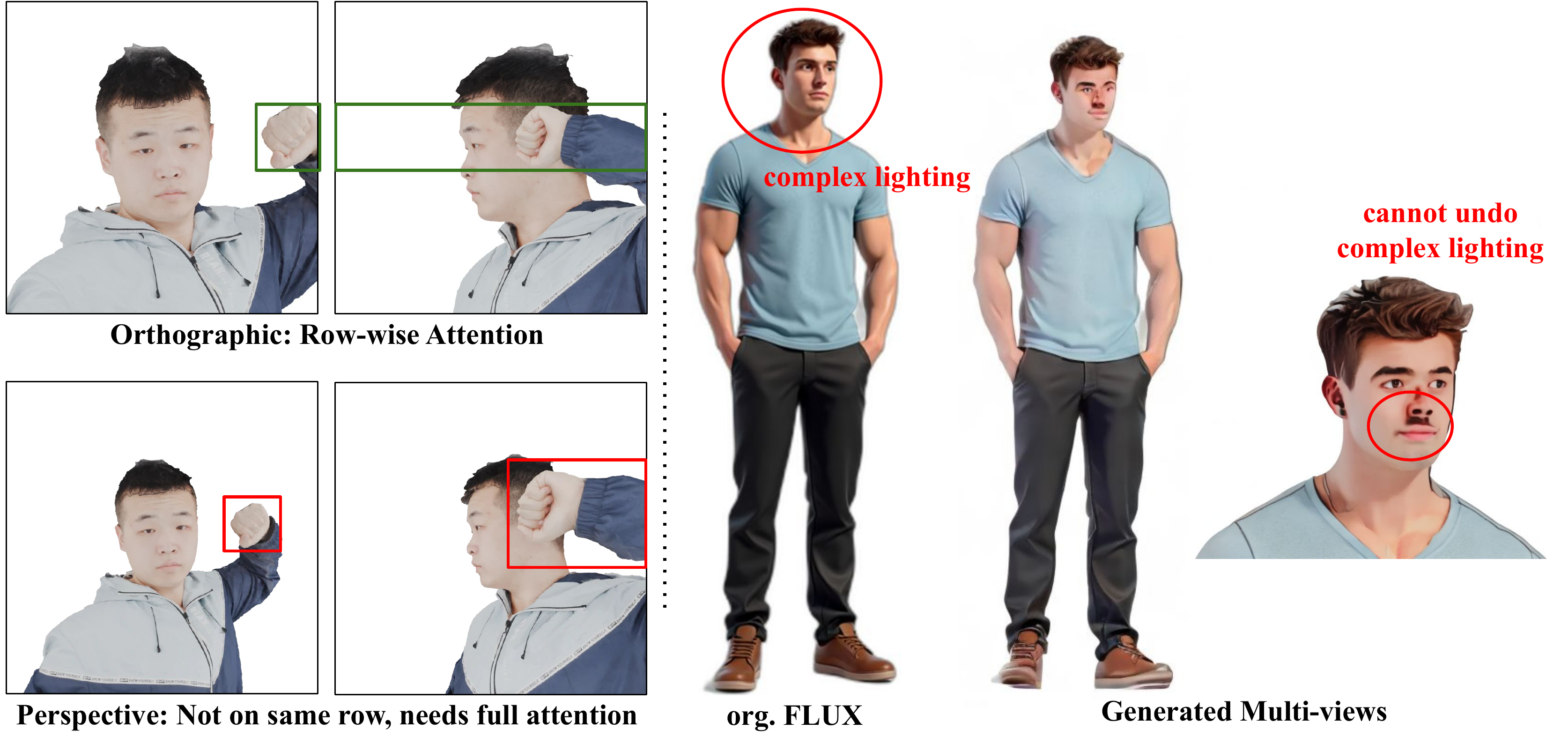}
    \captionof{figure}{Orthographic and Perspective in Multi-View Attention (left). Org. FLUX gives complex lighting, degrading multi-view generation (right).}
    \label{fig:ablation_multiview}
  \end{minipage}%
  \hfill
  \begin{minipage}{0.48\linewidth}
    \centering
    \includegraphics[width=\linewidth]{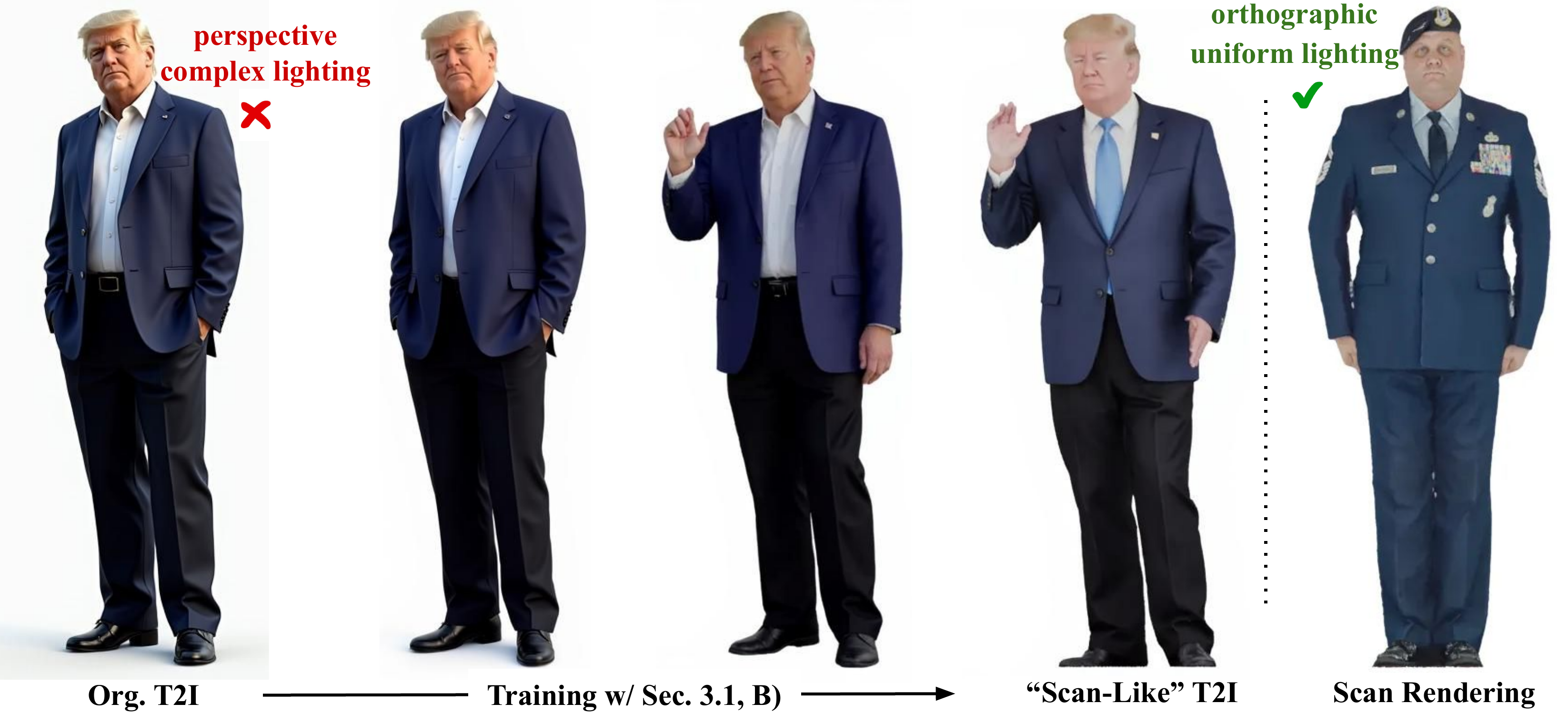}
    \captionof{figure}{Our finetuned FLUX can generate desired images from text prompt with orthographic view and uniform lighting, similar to the scan rendering.}
    \label{fig:ablation_desired_imgs}
  \end{minipage}

\end{figure*}

\clearpage
\newpage

\bibliographystyle{ACM-Reference-Format}
\bibliography{literatures}

\newpage

\begin{appendix}
\begin{figure*}
    \centering
    \includegraphics[width=0.95\linewidth]{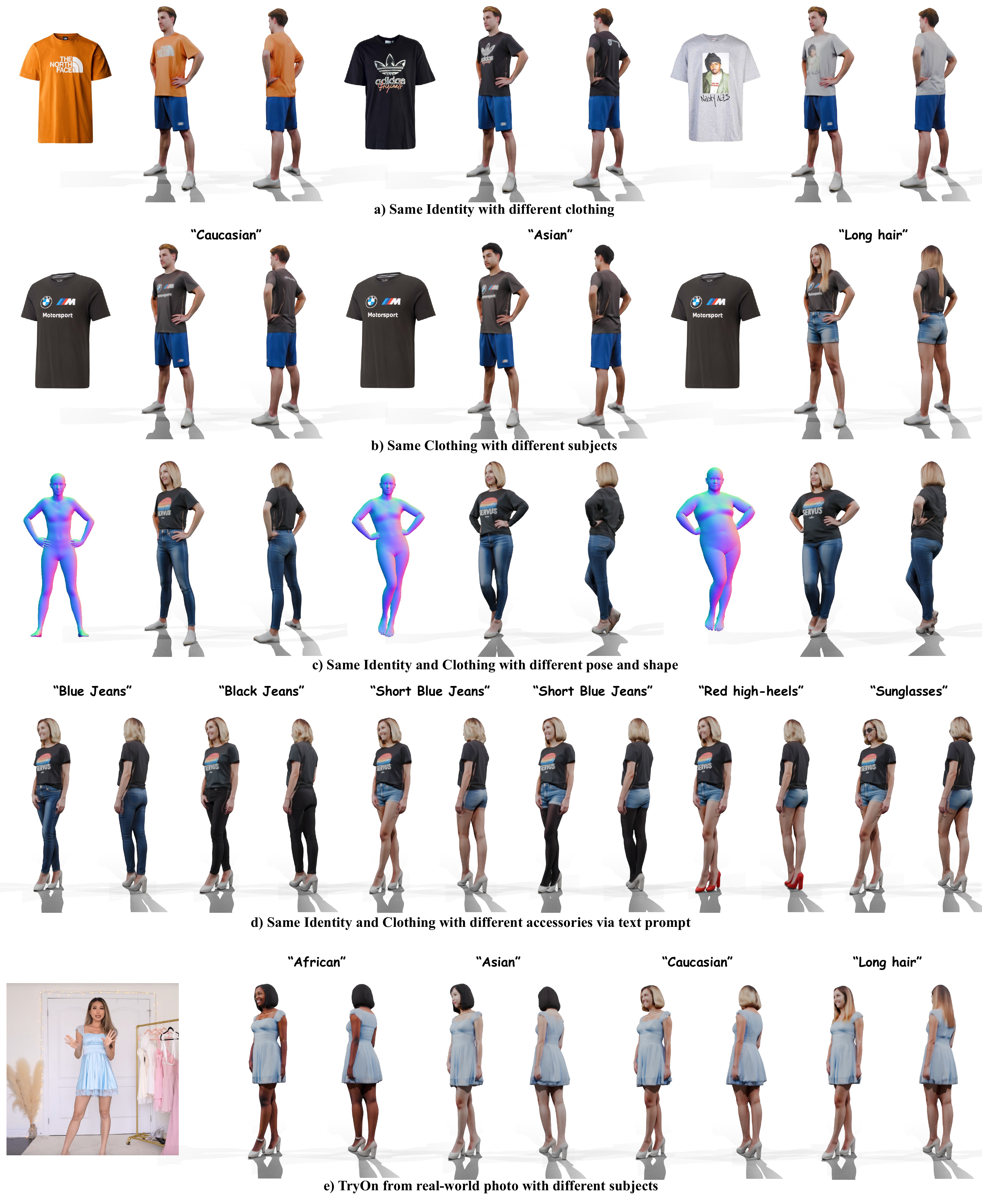}
    \caption{\textbf{Precise control capability in InfiniHumanGen}.}
    \label{fig:supp_infinihumangen}
\end{figure*}

\begin{figure*}
    \centering
    \includegraphics[width=0.95\linewidth]{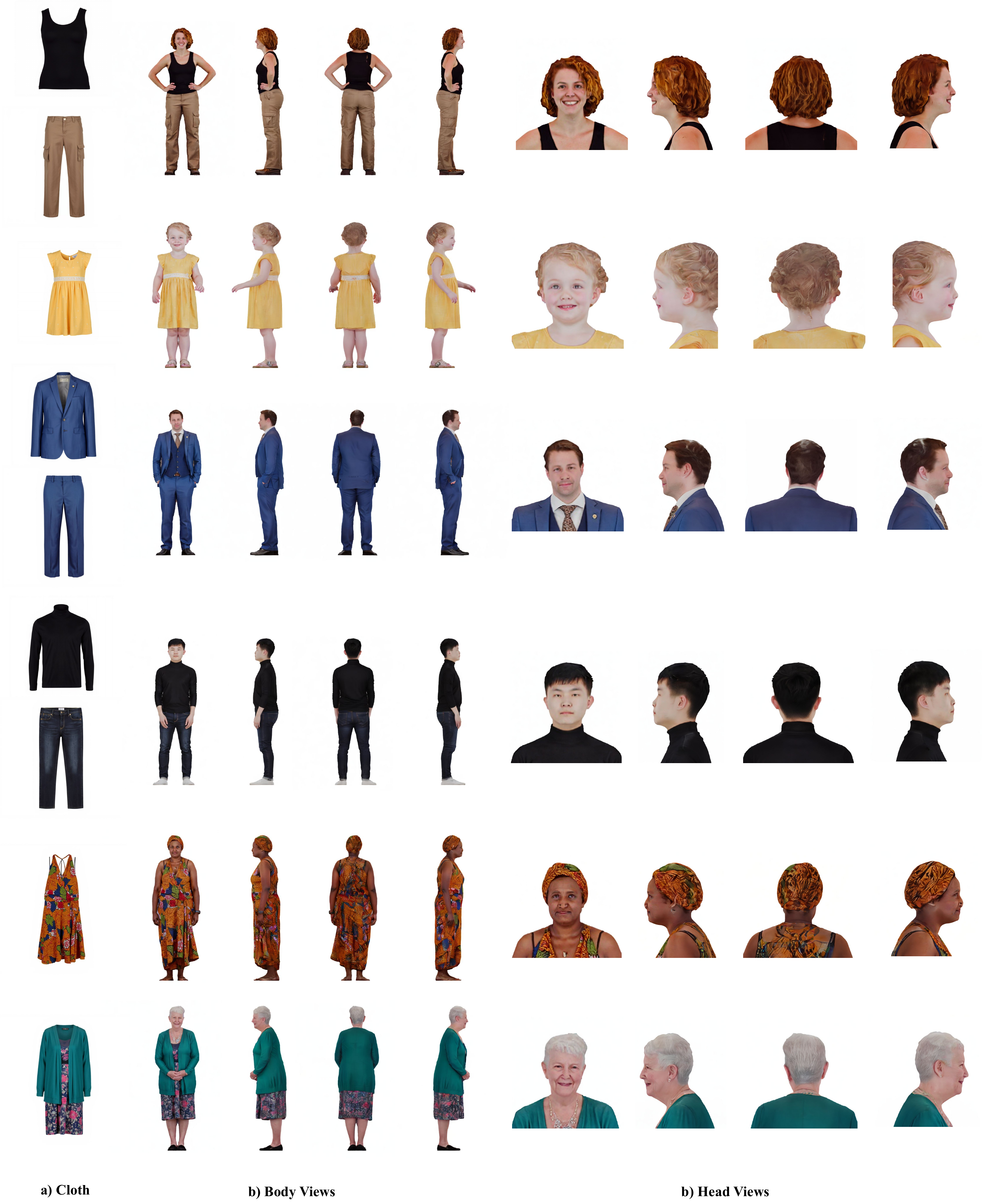}
    \caption{\textbf{Additional Examples in InfiniHumanData}. Our orthographic multi-view diffusion model generates high-quality view-consistent images of head views and full-body view.}
    \label{fig:supp_infinihumandata}
\end{figure*}

\begin{figure*}
    \centering
    \includegraphics[width=0.95\linewidth]{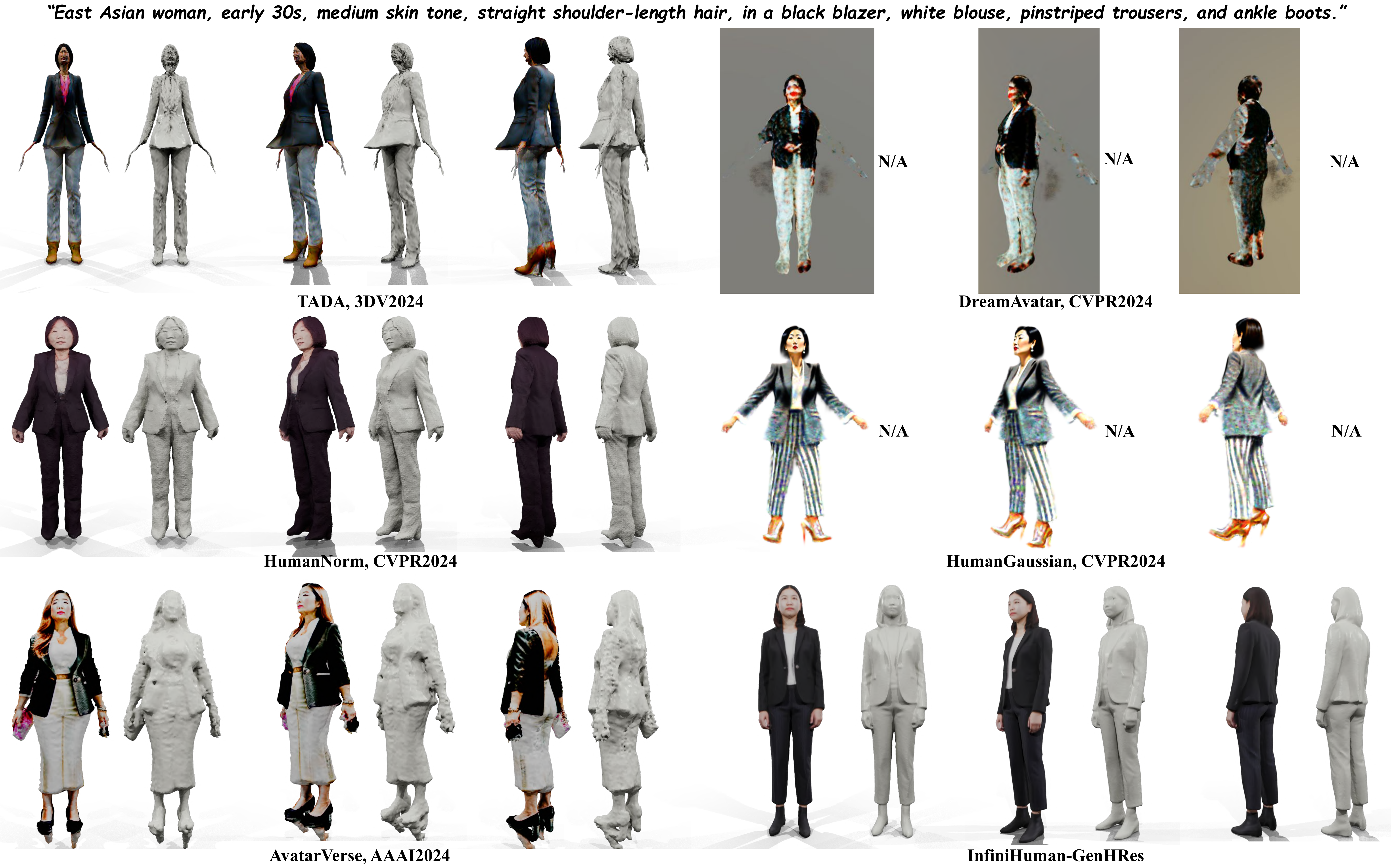}
    \includegraphics[width=0.95\linewidth]{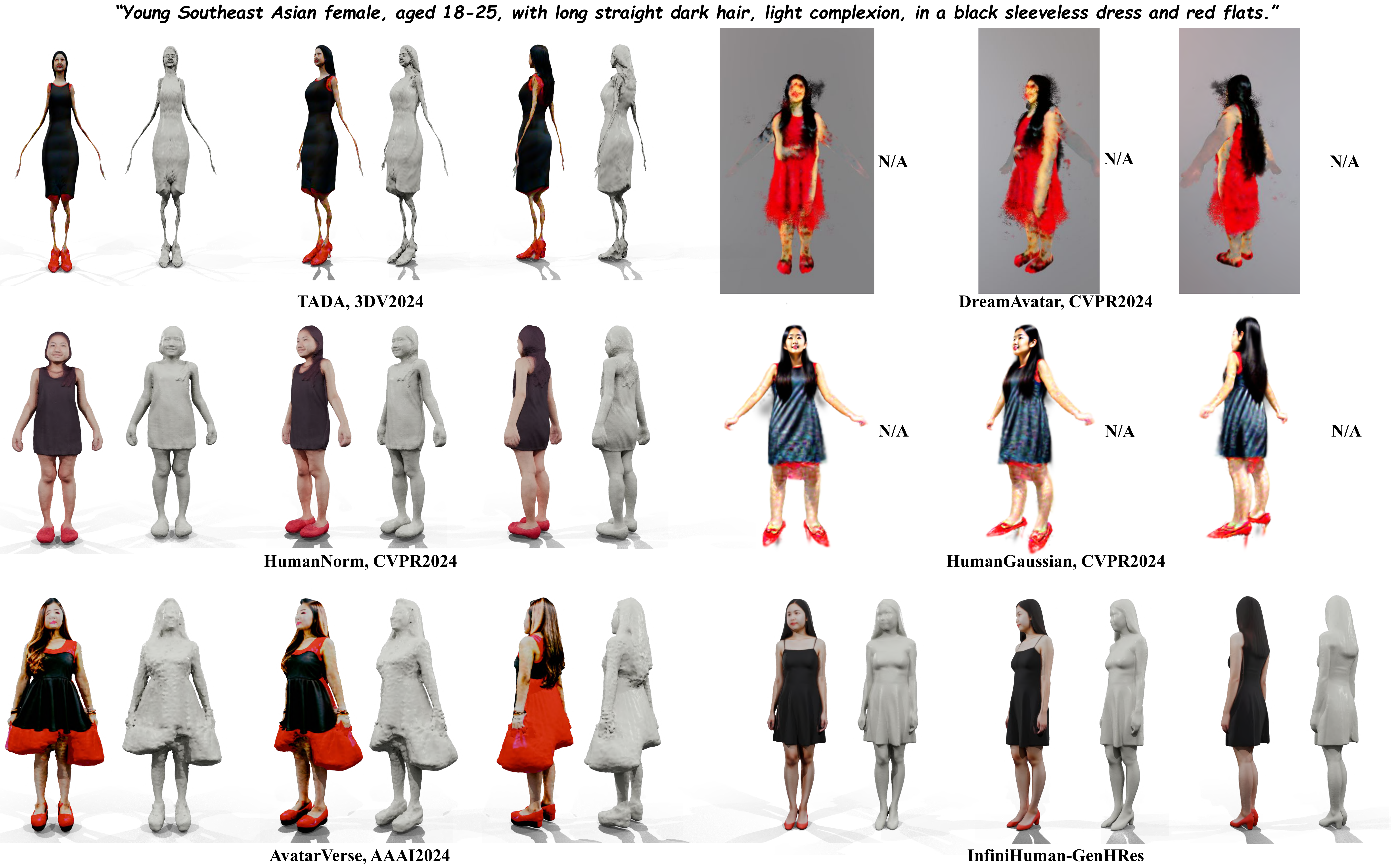}
    \caption{\textbf{Qualitative appearance and geometry comparison to SDS-based text-to-3D avatar approaches}.}
    \label{fig:supp_comparison_sds}
\end{figure*}

\begin{figure*}
    \centering
    \includegraphics[width=0.95\linewidth]{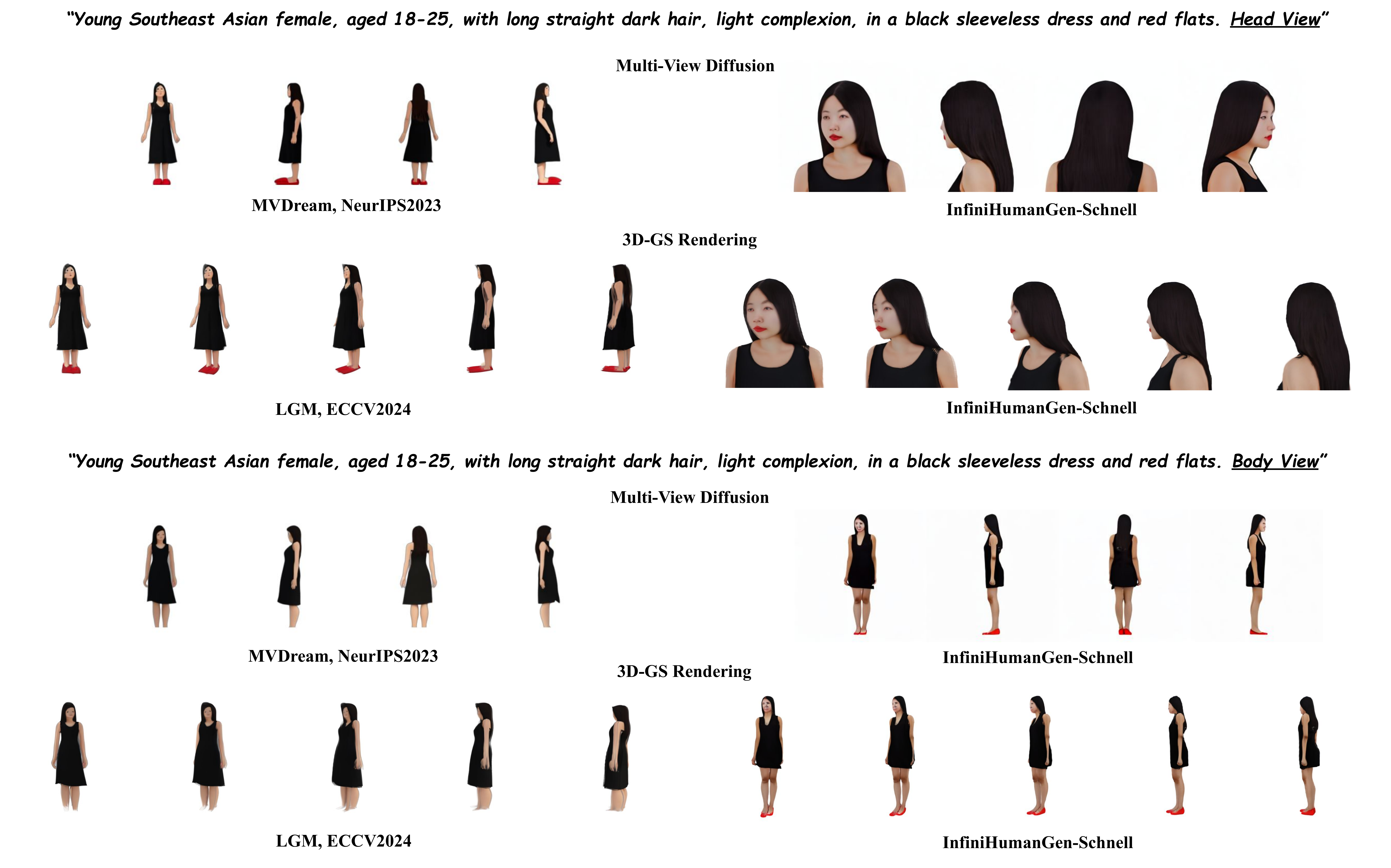}
    \includegraphics[width=0.95\linewidth]{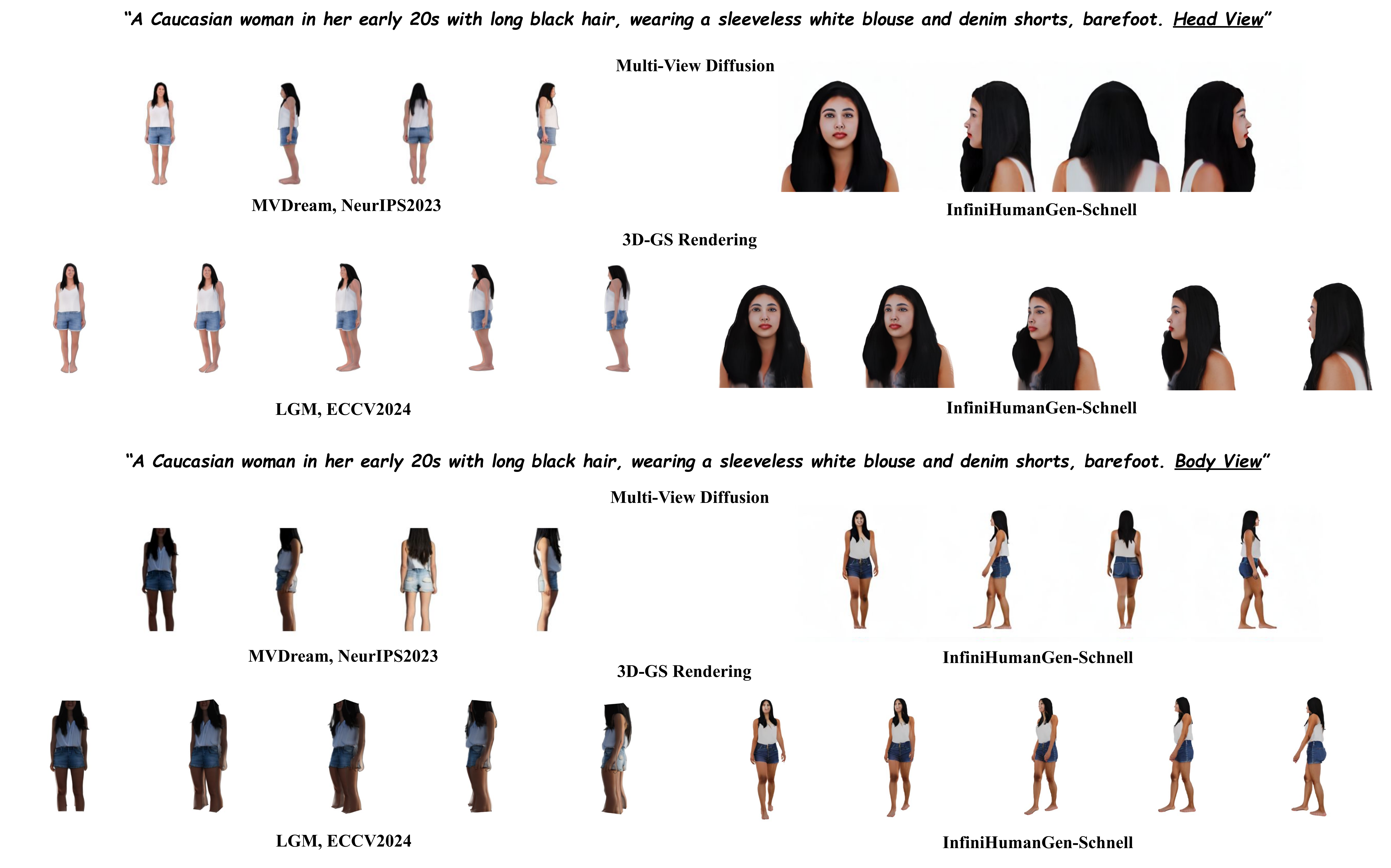}
    \caption{\textbf{Qualitative appearance and geometry comparison to feed-forward text-to-3D avatar approaches}.}
    \label{fig:supp_comparison_feedforward}
\end{figure*}

\begin{figure*}
    \centering
    \includegraphics[width=0.97\linewidth]{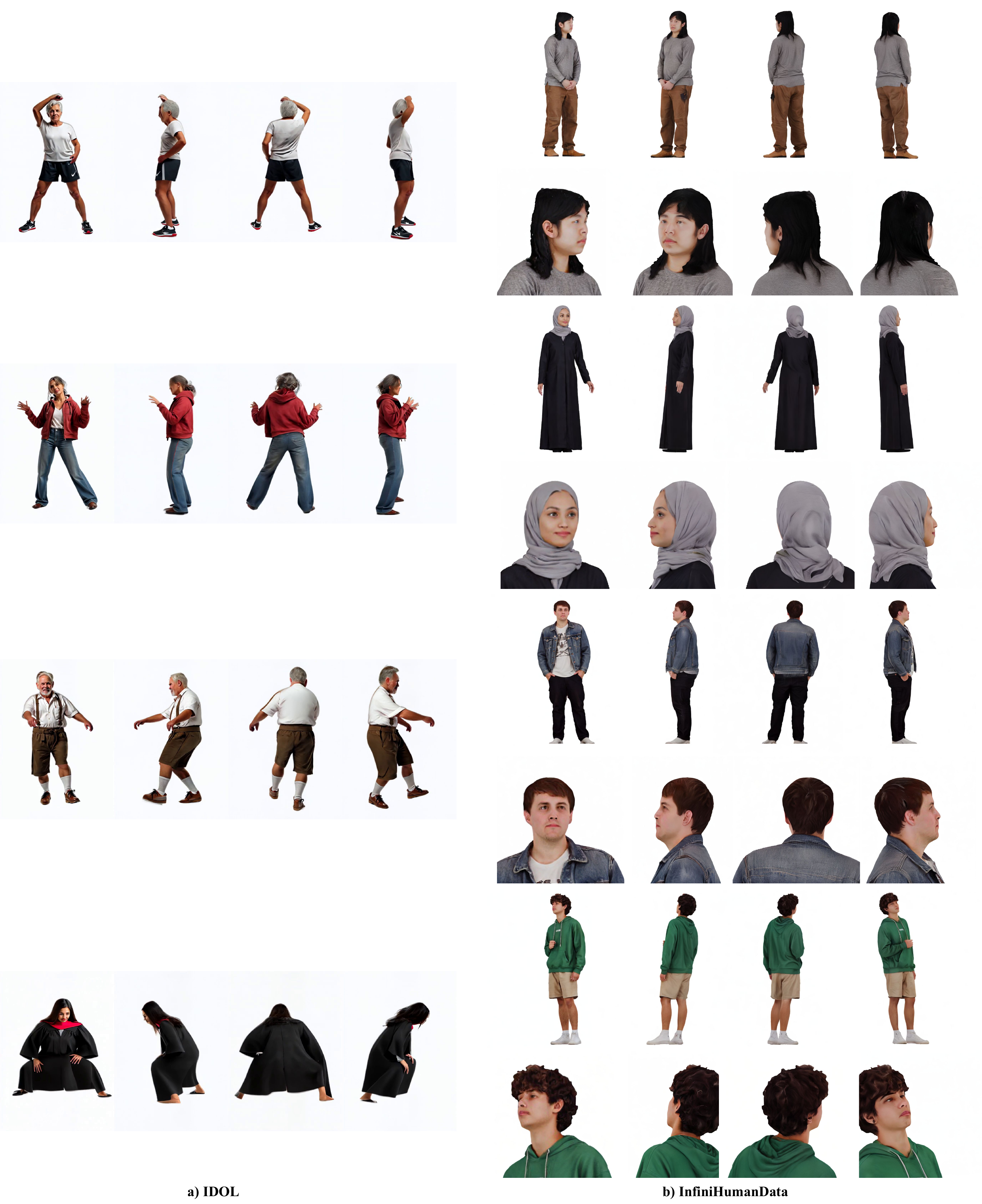}
    \caption{\textbf{Qualitative comparison with IDOL Dataset}. Our InfiniHumanData achieves better visual realism and multi-view consistency.}
    \label{fig:supp_comparison_idol}
\end{figure*}

\begin{figure*}
    \centering
    \includegraphics[width=\linewidth]{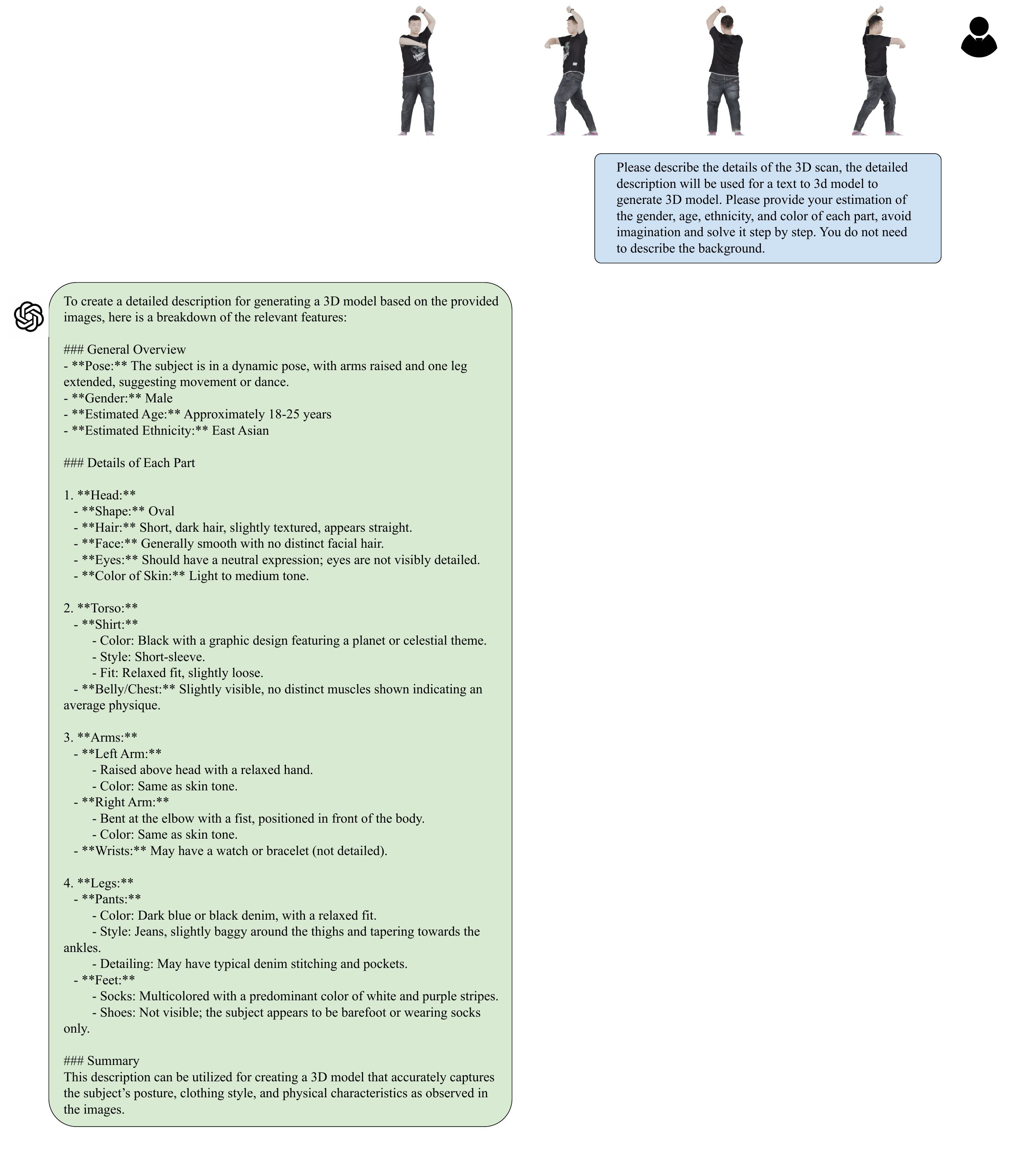}
    \caption{\textbf{ Detailed scan captioning prompt and example output}. }
    \label{fig:supp_scan_caption}
\end{figure*}

\begin{figure*}
    \centering
    \includegraphics[width=\linewidth]{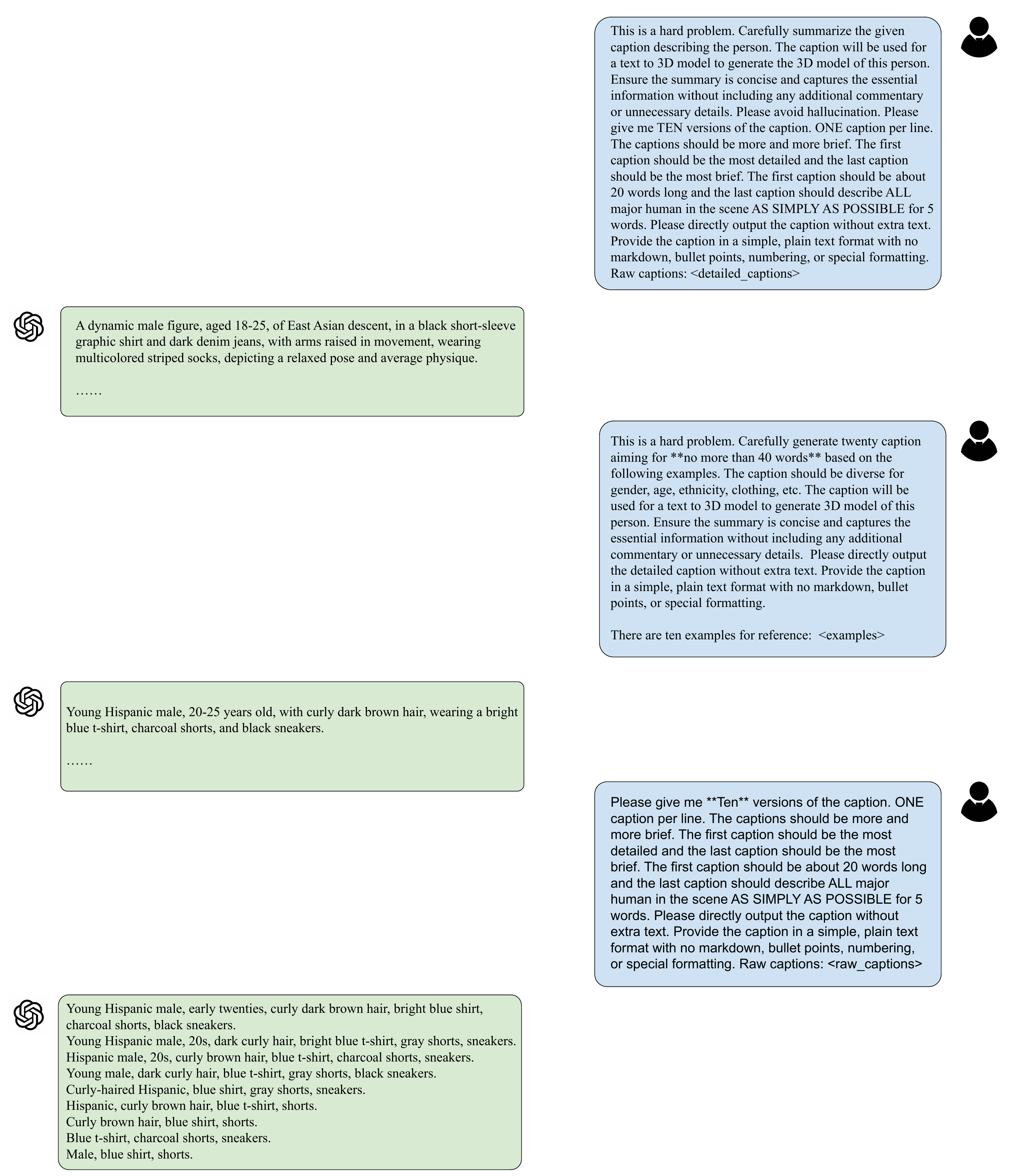}
    \caption{\textbf{Multi-granularity text summarization prompt and generated examples}. }
    \label{fig:supp_multi-granularity}
\end{figure*}

\begin{figure*}
    \centering
    \includegraphics[width=\linewidth]{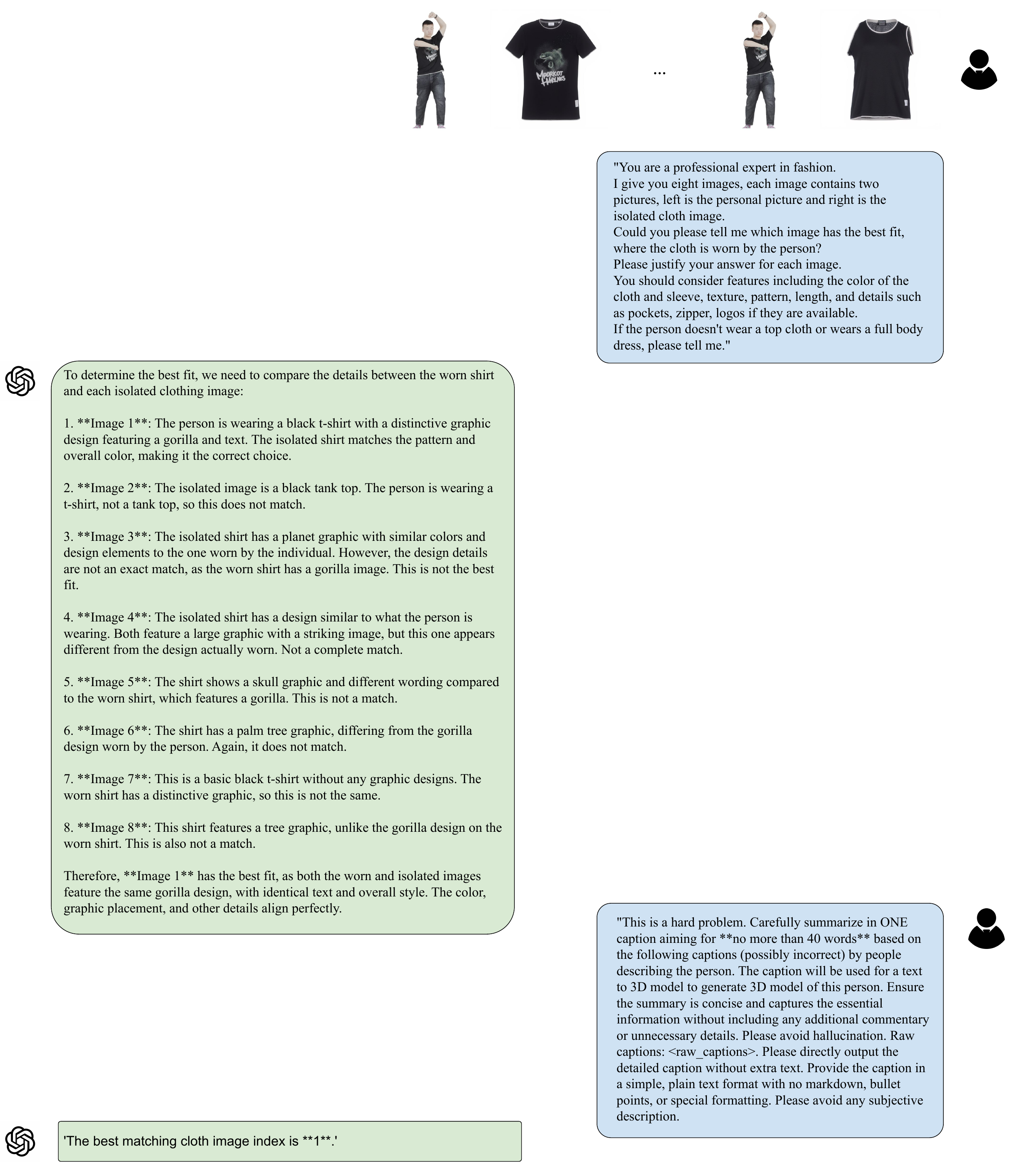}
    \caption{\textbf{Garment selection prompt for negative samples rejection.} }
    \label{fig:supp_negative_samples}
\end{figure*}

\begin{figure*}
    \centering
    \includegraphics[width=\linewidth]{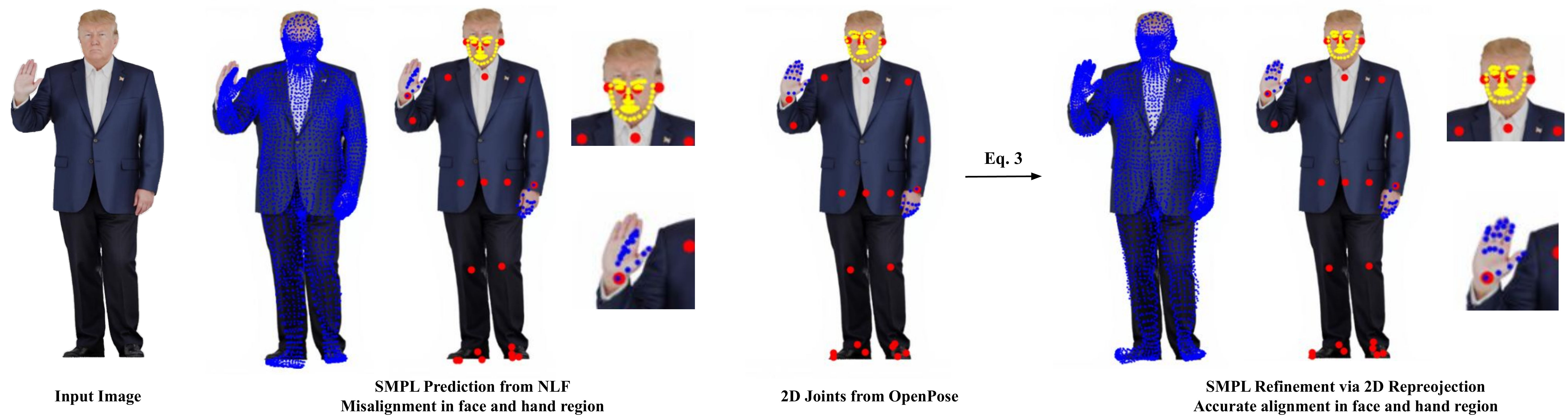}
    \caption{\textbf{Pose refinement using reprojection loss w.r.t. OpenPose 2D joints.} As stated in Sec.3.1 E in main paper, we use OpenPose 2D joints to further refine body pose parameters using reprojection loss. This additional operation eliminates the misalignment between image and SMPL body, and enhances the pixel-level accuracy in face and hand region.}
    \label{fig:supp_ablation_openpose}
\end{figure*}

\begin{figure*}
    \centering
    \includegraphics[width=\linewidth]{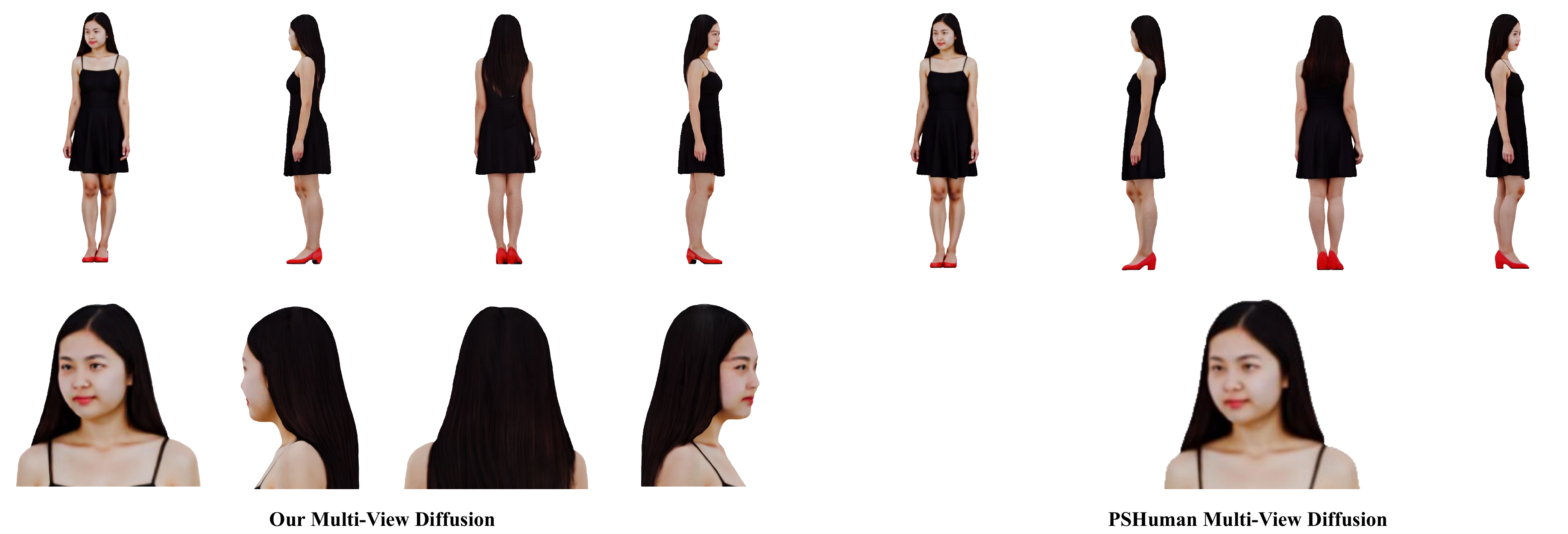}
    \caption{\textbf{Comparison between generated multi-view images in our model and PSHuman.} Both models generate high-resolution full-body images. However, our model generates multi-view high-resolution head-centric images, while PSHuman only generates a single blurry head-view image.}
    \label{fig:comparison_mvd_pshuman}
\end{figure*}

\begin{figure}
    \centering
    \includegraphics[width=0.9\linewidth]{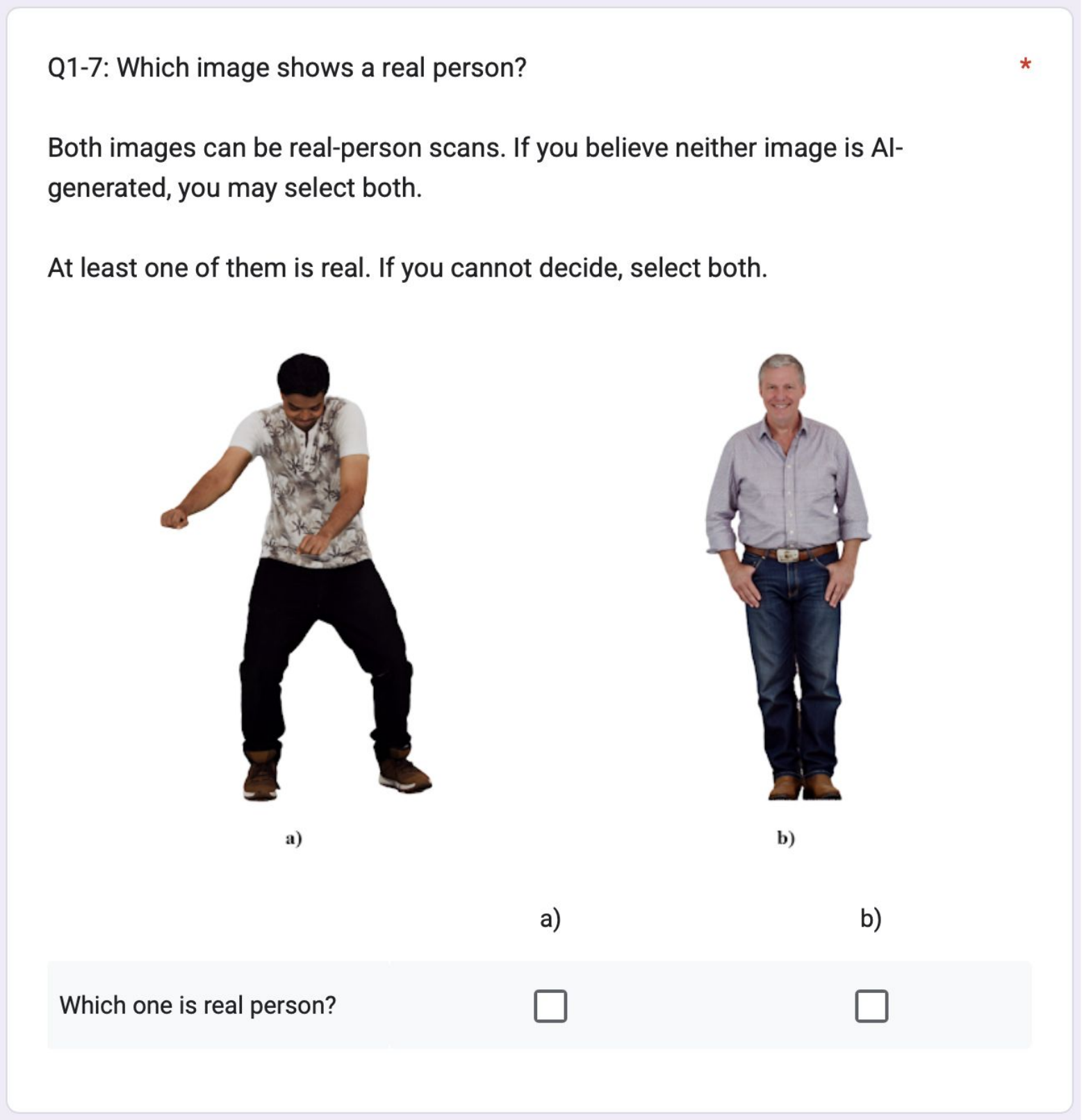}
    \caption{\textbf{User study for realism assessment of InfiniHumanData.} We present randomly paired InfiniHumanData subjects and scan subjects to participants. InfiniHumanData receives 765 votes over 746 votes for scan, which demonstrates the superior realism of our synthetic dataset.}
    \label{fig:userstudy_realism}
    \vspace{-1em}
\end{figure}

\section{Additional Visualitation}

\subsection{InfiniHumanGen}
Fig. ~\ref{fig:supp_infinihumangen} showcases the precise control capabilities of InfiniHumanGen. Our pipeline supports a wide variety of fine-grained controls, including changing identity, clothing, body shape, pose, and accessories via text, SMPL shape / pose, and clothing image inputs. As shown, we can manipulate specific attributes independently, such as varying clothing while maintaining identity, or editing pose and body shape with consistent appearance.

\subsection{InfiniHumanData}
In Fig. ~\ref{fig:supp_infinihumandata}, we present additional examples of InfiniHumanData subjects and their multi-modal annotations, including clothing assets and RGB multi-view images of head / body views. The figure highlights the diversity and annotation richness of our dataset across multiple attributes. The visualization of InfiniHumanData demonstrates our orthographic multi-view diffusion model generates view-consistent images for head views and full-body views.

\subsection{InfiniHumanGen vs. SOTA}
We present additional qualitative comparisons between InfiniHumanGen and state-of-the-art (SOTA) text-to-3D avatar generation methods. Fig. ~\ref{fig:supp_comparison_sds} compares our results against prominent SDS-based approaches~\cite{zhang2023avatarverse, cao2023dreamavatar, liao2023tada, huang2024humannorm, liu2024humangaussian}, while Fig. ~\ref{fig:supp_comparison_feedforward} shows results against recent feed-forward pipelines~\cite{shi2024mvdream, tang2024lgm}.

Compared to SDS-based methods, such as DreamFusion and its variants, InfiniHumanGen consistently produces avatars with significantly higher visual fidelity, more consistent geometry, and superior alignment with the input prompt. SDS-based methods often suffer from slow convergence and visual artifacts such as over-smoothed textures or structural inaccuracies, whereas our pipeline delivers sharper appearance details and well-formed geometry.

We additionally compare with Chupa~\cite{kim2023chupa}, a human mesh generation approach controllable via text prompts and SMPL pose. As shown in Fig., although Chupa accurately follows the pose, it fails to generalize to complex text prompts. Moreover, it does not support the use of specific clothing images as conditioning input. In contrast, our InfiniHuman-Gen can seamlessly handle complex text prompts and the additional clothing image.

When compared to feed-forward approaches, InfiniHumanGen demonstrates stronger text-following capability, more accurate pose and clothing reproduction, and improved overall realism. Competing methods frequently show issues such as geometry distortion, incorrect clothing assignment, or poor text-following ability for generating head avatars. In contrast, our pipeline generates view-consistent avatars that faithfully reflect the user-provided conditions.

For all examples, InfiniHumanGen offers rapid generation speed, fine-grained attribute control, and robust multi-modal conditioning, setting a new standard for controllable 3D human avatar synthesis.

\subsection{InfiniHumanData vs. SOTA}
To further highlight the advantages of InfiniHumanData over previous large-scale human datasets, we provide a visual qualitative comparison with the recent IDOL~\cite{Zhuang2025idol} dataset. While IDOL represents a significant step forward in generating multi-view human images from single inputs using video diffusion models, we observe that its generated results often exhibit noticeable view inconsistencies and temporal artifacts. This is primarily due to the neighbor-only attention mechanism used in video diffusion, which can lead to unnatural transitions and inconsistent appearance across views.

In contrast, InfiniHumanData leverages multi-view diffusion and a carefully designed generation pipeline to ensure high-resolution, view-consistent, and photorealistic renderings for each identity. Our approach produces multi-view images that maintain consistent shape, texture, and lighting, closely resembling the outputs of true 3D scan renderings.

We present several side-by-side examples comparing multi-view images from both datasets in Fig.~\ref{fig:supp_comparison_idol}. As shown in the figure, InfiniHumanData consistently produces sharper details, smoother transitions, and significantly improved cross-view coherence compared to IDOL. The comparison demonstrates our multi-view diffusion model achieves more view-consistent results than video-diffusion models in IDOL. These results reinforce the value of our method for downstream tasks that require highly realistic and consistent human data, such as 3D reconstruction, animation, and virtual try-on.

\section{InfiniHumanData Implementation Details}

\subsection{Scan Captioning}
We generate scan captions using a two-step process with GPT-4o~\cite{gpt4o2024}. First, a structured prompt (see Fig.~\ref{fig:supp_scan_caption}) guides the model to extract key attributes—gender, age, ethnicity, pose, clothing, and body parts—directly from scan images, minimizing hallucination and irrelevant details. This approach ensures each caption concisely summarizes visual appearance and clothing for downstream tasks. The same prompt is used for all samples; example outputs and the full prompt are in Fig.~\ref{fig:supp_scan_caption}.

\subsection{Multi-granularity Text Generation}
To support flexible conditioning, we expand each detailed caption with multiple text granularities. Using a summarization prompt (Fig.~\ref{fig:supp_multi-granularity}), GPT-4o generates a sequence of increasingly concise captions, from detailed (about 20 words) to minimal (5 words). This enables models to learn from both fine-grained and coarse cues and improves robustness to user input. All examples use the prompt in Fig.~\ref{fig:supp_multi-granularity}.

\subsection{Negative Samples Rejection}
To ensure garment accuracy, we generate several clothing candidates per subject and use GPT-4o with a comparison prompt (Fig.~\ref{fig:supp_negative_samples}) to select the best match. The model compares candidates based on color, texture, pattern, length, and fine details (e.g., pockets, zippers), justifying its choice. If none match, the model returns ‘No’. Full prompt and an example are shown in Fig.~\ref{fig:supp_negative_samples}.

\subsection{Monocular Body Fitting}
As stated in Sec. 3.1 E of the main paper, we use a two-stage approach to obtain the SMPL pose and shape parameters for InfiniHumanData. Although NLF~\cite{sarandi2024nlf} can estimate accurate shape parameter, the estimated pose still produces misalignment of the SMPL body and the images. Hence, we use estimated 2D joints from OpenPose~\cite{cao2019openpose} to refine the SMPL pose parameters with the reprojection error. As illustrated in Fig.~\ref{fig:supp_ablation_openpose}, the additional 2D joints-based refinement successfully resolves the issue and ensures the pixel-level alignment of the body parameters and the images, especially in face and hand region. This alignment is crucial in the SMPL-guided multi-view generation, which can be found in Fig.~13 of the main paper.

\section{InfiniHumanGen Implementation Details}
\subsection{Gen-Schnell.} 
\noindent\textbf{Fine-tuning MVDream.} 
Since MVDream is originally designed to condition only on text and does not support additional modalities such as SMPL normal maps or clothing images, we slightly adapt its structure while preserving the power of the pretrained model. Specifically, we expand the input convolutional layer of the UNet denoiser from 4 channels to 12 channels, allowing us to concatenate the VAE-encoded SMPL normal map and clothing image with the initial VAE Gaussian noise as input.

To ensure that the introduction of these new channels does not disrupt the pretrained behavior, we initialize the weights associated with the additional channels to zero. Formally, let $\mathbf{x}_0 \in \mathbb{R}^{B \times 4 \times H \times W}$ denote the original input, and $\mathbf{x}'_0 \in \mathbb{R}^{B \times 12 \times H \times W}$ the augmented input. The new input convolution weights $\mathbf{W}' \in \mathbb{R}^{C \times 12 \times k \times k}$ are set as:
\begin{equation}
    \mathbf{W}' = \begin{bmatrix} \mathbf{W} & \mathbf{0} \end{bmatrix}
    \label{eq:zero_init_weights}
\end{equation}
where $\mathbf{W} \in \mathbb{R}^{C \times 4 \times k \times k}$ are the pretrained weights, and $\mathbf{0}$ represents zero-initialized weights for the new channels.

As a result, the output of the modified convolution at initialization is unchanged for any input where the additional channels are zero, thus strictly preserving the pretrained functionality:
\begin{equation}
    \text{Conv}(\mathbf{x}'_0; \mathbf{W}') = \text{Conv}(\mathbf{x}_0; \mathbf{W}) \quad \text{when} \quad \mathbf{x}'_0 = [\mathbf{x}_0, \mathbf{0}]
    \label{eq:zero_init_equiv}
\end{equation}

During finetuning, the model can gradually learn to utilize the new SMPL and clothing conditions to improve generation.

\noindent\textbf{Training.}
After fine-tuning MVDream on InfiniHumanData, the resulting 2D diffusion models demonstrate strong multi-view image generation capabilities conditioned on text, SMPL normal maps, and clothing images. To leverage this for 3D reconstruction, we train the Splat Decoder separately on 6{,}000 high-quality human scans. For each scan, we render multi-view images and use them as supervisory signals, which significantly improves convergence and stability in the prediction of 3D Gaussian Splat representations. 
We adopt similar strategy for diffusing 2D multi-view images and the 3D Gaussian Splat in Human-3Difusion~\cite{xue2024human3diffusion}. Apart from different conditioning of text, the main difference here is that our fine-tuned multi-view diffusion model generates orthographic images which is not directly align with perspective images rendered by 3D Gaussian splatting. Thus, we only apply the consistent reverse sampling for beginning 10 DDIM steps. It allows the 2D orthographic multi-view diffusion and 3D Gaussian diffusion synchronize closely, but the final generation is not confused by the inconsistency of the orthographic or perspective generation. Please refer to our implementation for more details.

During training, we use an Adam optimizer with an initial learning rate of $5 \times 10^{-4}$ and a batch size of 256. The model is trained on 8 NVIDIA A100 GPUs. To further stabilize training, we apply spectral normalization to the decoder and use a perceptual loss (LPIPS) in addition to $\ell_2$ reconstruction loss on multi-view images.

\subsection{Gen-HRes.}
We select OminiControl2~\cite{tan2025ominicontrol2} as our backbone for multi-conditional image-to-image translation, and fine-tune it on InfiniHumanData for our application. Specifically, given a SMPL body normal map, a clothing image, and a detailed text description, our goal is to generate a full-body image that faithfully reflects all three input modalities.

In the original OminiControl2 framework, input image conditions are categorized as either spatially aligned (e.g., depth maps, normal maps, canny edges) or non-aligned (e.g., object images). In our design, the SMPL normal map serves as a spatially aligned condition, as it conveys approximate pixel-wise spatial information about the generated avatar. Thus, we assign it a dynamic positioning token of 0. For the clothing image, which provides semantic information but lacks pixel-level alignment, we treat it as a non-aligned condition and assign a dynamic positioning token of -48. This design enables OminiControl2 to appropriately fuse information from both modalities and condition the image generation process accordingly. For further architectural details, please refer to the original OminiControl2 paper.

For the multi-modal image generator in InfiniHuman-GenHRes, we use the Flux-Dev~\cite{flux2024} as the base model and train LoRA to accommodate additional multimodal cloth image and input body shape. The training resolution is set to 768$\times$ 768, and the LoRA rank is set to 16. Our model is trained with a batch size of 1 and gradient accumulation of 8 steps. We employ the Prodigy optimizer with safeguard warm-up and bias correction enabled, setting the weight decay to 0.01. The model is trained for 35000 steps.

Our InfiniHuman-GenHRes model leverages the fine-tuned orthographic multi-view diffusion model to lift the generated single images to multi-views, similar to PSHuman~\cite{li2024pshuman}. However, PSHuman only generates a low-resolution single head view and resized to high-resolution, while our model generates directly multi-view high-resolution head images. This design allows us to obtain more details and fidelity of the generated heads such as eyes, ears, and hair.

\end{appendix}

\newpage

\end{document}